%% file: template.tex
\documentclass{article}

\usepackage{arxiv}

\usepackage[utf8]{inputenc} % allow utf-8 input
\usepackage[T1]{fontenc}    % use 8-bit T1 fonts
\usepackage{hyperref}       % hyperlinks
\usepackage{url}            % simple URL typesetting
\usepackage{booktabs}       % professional-quality tables
\usepackage{amsfonts}       % blackboard math symbols
\usepackage{nicefrac}       % compact symbols for 1/2, etc.
\usepackage{microtype}      % microtypography
\usepackage{lipsum}		% Can be removed after putting your text content
\usepackage{graphicx}
\usepackage[numbers]{natbib}
\usepackage{doi}
\usepackage{amsmath}
\usepackage{xcolor} % Required for color support
\usepackage{soul}   % Provides the \hl command
\usepackage{gensymb}
\usepackage{tabularx}
\usepackage{changepage}
\usepackage{amssymb}
\usepackage{multirow}
\usepackage{float}
\usepackage{bbm}
\title{STA-TFM: Spatio-Temporal Aggregation Across Views TransForMer for Pose Estimation}

\author{ 
    \href{https://orcid.org/0009-0007-1380-6950}{Mena Kamel}$^{\dagger,*}$ \\
	Sanofi Digital R\&D, \\
    240 Richmond St. W, Toronto, ON \\
    M5V 1V6, Canada \\
	\And
    Natalie Won$^{\dagger}$ \\
	Sanofi Digital R\&D, \\
    240 Richmond St. W, Toronto, ON \\
    M5V 1V6, Canada \\
    \And
    Amrut Sarangi \\
	Sanofi Digital R\&D, \\
    450 Water St, Cambridge, MA \\
    02141, USA \\
    \And
    Sven Jager \\
	Sanofi Digital R\&D, \\
    K703, Industriepark Höchst, \\
    65929 Frankfurt am Main, Germany \\
    \And
    \href{https://orcid.org/0000-0001-7527-2500}{Albert Pla Planas}$^{*}$ \\
	Sanofi Digital R\&D, \\
    Carrer de Rosselló i Pòrcel, 21, \\
    Nou Barris, 08016 Barcelona, Spain
}

\renewcommand{\shorttitle}{STA-TFM: Multi-View 3D Pose Estimation}
\newlength{\extralength}
\let\oldtabularx\tabularx
\renewcommand{\tabularx}[2]{\small\oldtabularx{#1}{#2}}
\newlength{\fulllength}
\newcolumntype{C}{>{\centering\arraybackslash}X}

\makeatletter
\newcolumntype{2}{>{\hsize=2\hsize}X}
\makeatother

\hypersetup{
pdftitle={STA-TFM: Spatio-Temporal Aggregation Across Views TransForMer for Pose Estimation},
pdfsubject={Computer Vision, Human Pose Estimation},
pdfauthor={Mena Kamel, Natalie Won, Amrut Sarangi, Sven Jager, Albert Pla Planas},
pdfkeywords={3D Human Pose Estimation, Spatio-Temporal Transformers, Multi-view Fusion},
colorlinks=true,
linkcolor=black,
citecolor=black,
urlcolor=black,
}

\begin{document}
\maketitle

% Add author notes AFTER \maketitle
\renewcommand{\thefootnote}{\fnsymbol{footnote}}
\footnotetext[2]{$^{\dagger}$ These authors contributed equally to this work.}
\footnotetext[0]{$^{*}$ Correspondence: \texttt{mena.kamel@sanofi.com}, \texttt{albert.plaplanas@sanofi.com}}
\renewcommand{\thefootnote}{\arabic{footnote}}

\begin{abstract}
Monocular 3D human pose estimation (HPE) remains challenging due to depth ambiguity, occlusions, and the need for temporal consistency. While multi-view methods provide superior accuracy over monocular approaches, they often require complex setups. We introduce STA-TFM, a transformer-based architecture that combines spatial and temporal information for multi-view pose estimation. The approach leverages DSTformer, a monocular feature extractor, to capture long-range pose dependencies within each view. A fusion transformer then aggregates information across views to produce coherent 3D estimates. To address training data scarcity, we use a data generation pipeline that transforms any existing 3D pose dataset into multi-view setups with controllable parameters. Experiments on various datasets demonstrate that STA-TFM outperforms existing camera-parameter-free multi-view methods. STA-TFM achieves 50.9\% and 49.5\% reductions in mean per joint position error (MPJPE) and mean per joint velocity error (MPJVE) on the DHP19 dataset. Furthermore, it achieves 6.7\% and 7.7\% respective reductions on HAA4D, and a 15.2\% MPJPE reduction on TotalCapture. STA-TFM handles noisy and missing 2D inputs, supporting potential deployment in healthcare monitoring, athletic assessment, and immersive technologies. Code, training checkpoints, and data are available at \url{https://zenodo.org/records/22832620}.
\end{abstract}

% keywords can be removed

\section{Introduction}
\input{MDPI_Article_Template/Sections/1_intro}
\section{Related Works}
\input{MDPI_Article_Template/Sections/2_related_works}
\section{Materials and Methods}
\input{MDPI_Article_Template/Sections/3_methodology}

\section{Results}
\input{MDPI_Article_Template/Sections/4_experiments}

\section{Discussion}
\input{MDPI_Article_Template/Sections/5_conclusion}

\section*{Data Availability}
Source code, training checkpoints, and sample data are publicly available at \url{https://zenodo.org/records/22832620}.

\section*{Acknowledgments}
 We thank the creators of the DHP19, HAA4D, BlendMimic3D, CMUPanoptic, and TotalCapture datasets for making their data publicly available, which enabled the comprehensive evaluation presented in this work. During the preparation of this manuscript, the authors used Claude for text refinement and manuscript editing. The authors also used GitHub Copilot (running Claude Sonnet 4.6 and Claude Opus 4.7) for code development tasks including code cleanup and debugging. The authors have reviewed and edited the output and take full responsibility for the content of this publication.

\section*{Conflicts of Interest}
% This work was fully funded by Sanofi Digital R\&D. 
All authors are Sanofi employees and may hold shares and/or stock options in the company.

\appendix
\section{Appendix}
\subsection{Extended Comparisons}
\input{MDPI_Article_Template/Sections/appendix}

\bibliographystyle{unsrtnat}
\bibliography{MDPI_Article_Template/Sections/references.bib}

\end{document}

%% file: MDPI_Article_Template/Sections/1_intro.tex
\label{sec:intro}

Human pose estimation from video is a fundamental computer vision challenge with broad applications spanning healthcare and athletics ~\cite{roggio2024comprehensive}, virtual reality and gaming ~\cite{caserman2019survey}, and beyond ~\cite{dibenedetto2025comparing,zheng2023deeplearningbasedhumanpose}, where developing accurate models holds significant promise. To obtain 3D estimates from videos, the task is often broken into two stages: 2D pose extraction followed by 2D to 3D lifting. While significant advancements have been made in 2D pose extraction from videos, obtaining 3D poses remains a complex problem ~\cite{zheng2023deeplearningbasedhumanpose}. Two methods have emerged to address lifting: monocular and multi-view human pose estimation (HPE), each with unique trade-offs between accuracy and accessibility. Monocular methods are more accessible as they rely on a single view, circumventing the need for complex and expensive data acquisition processes ~\cite{guo2025survey}. However, the single view results in an ill-posed problem in which a single 2D observation represents multiple 3D locations ~\cite{kim2024mhcanonnet}. Depth ambiguity becomes more challenging in the presence of external and self-occlusions~\cite{guo2025survey}. While monocular pose estimation is typically easier to deploy, it is often at the cost of accuracy. Without accurate 3D estimation, joint angles and movement patterns cannot be relied on to inform motion assessments. These challenges motivate the implementation of multi-view setups where 2D to 3D lifting relies on multiple cameras, making these models robust and better candidates for real-world applications. However, multi-view models often require complex and expensive setups for data acquisition, making them poor candidates for `in-the-wild' scenarios and practical day-to-day monitoring ~\cite{zheng2023deeplearningbasedhumanpose, kim2024mhcanonnet}. 

In monocular and multi-view methods, transformer-based architectures have gained traction in recent years for their ability to model long-range spatial and temporal dependencies ~\cite{zhao2023poseformerv2}. Monocular models that harness spatial and temporal information demonstrate that attention mechanisms can improve temporal coherence and pose accuracy ~\cite{zhu2023motionbert,mehraban2024motionagformer}. These benefits extend naturally to multi-view settings, where synchronized information from multiple cameras can be leveraged to obtain more robust 3D estimates. Existing methods exemplify this by fusing per-view spatial embeddings with a fusion pose transformer (FPT) to obtain reliable 3D poses from multiple 2D poses ~\cite{ghasemzadeh2024mpl}.

\begin{figure}[H]
%\isPreprints{\centering}{} % Only used for preprints
\centering
\includegraphics[width=8.0cm]{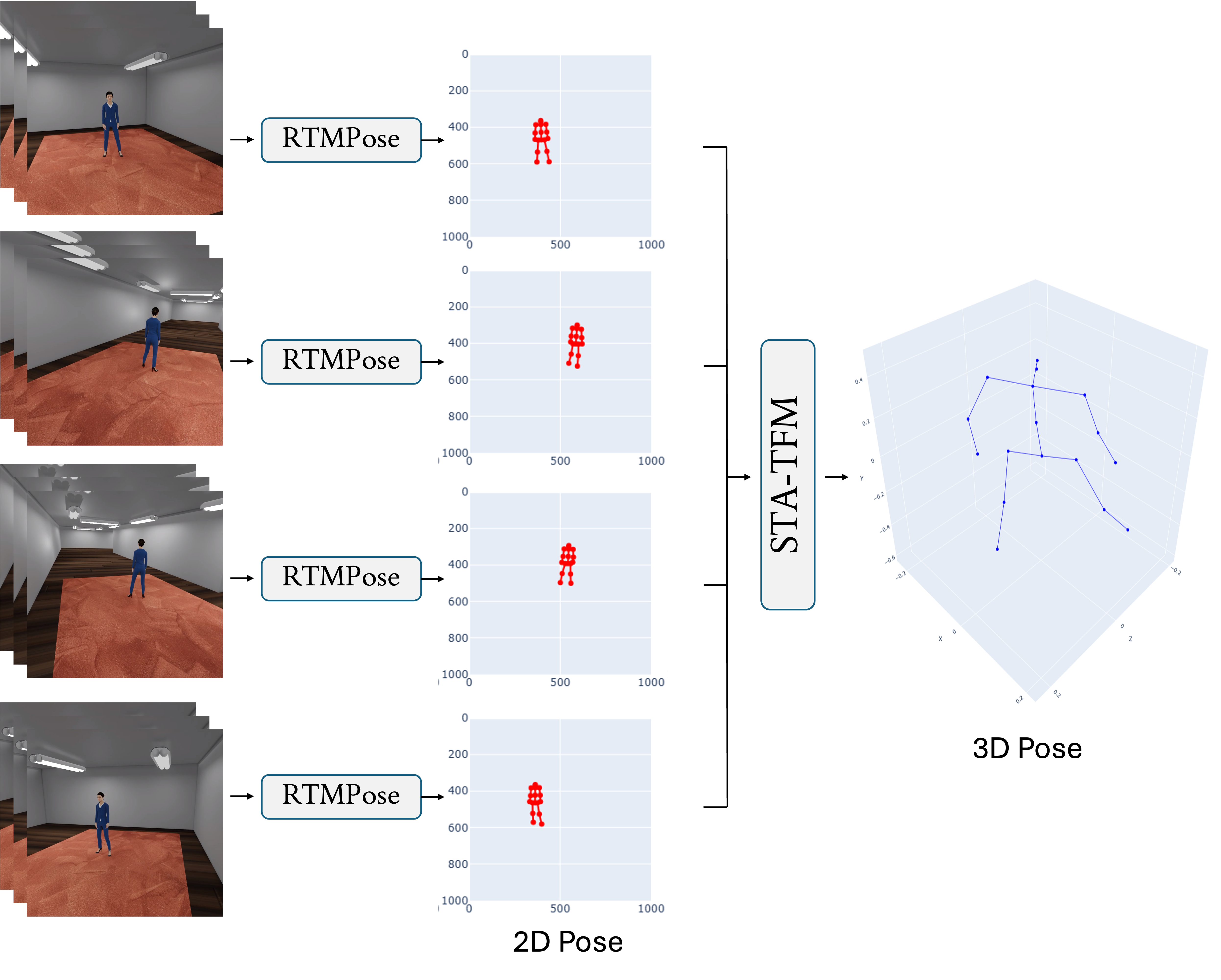}
\caption{Starting with multi-view videos, a 2D pose estimation model (e.g., RTMPose \cite{jiang2023rtmpose}) extracts 2D joint locations from each camera view. These multi-view 2D poses are then processed by STA-TFM through spatio-temporal aggregation to predict the corresponding 3D pose. Only the videos are required; camera calibration parameters are not needed during inference. Example shown is from the BlendMimic dataset \cite{lino20243d}.\label{fig:sta_tfm_pipeline}}
\end{figure}

Building on transformer-based multi-view methods, we introduce STA-TFM (\textbf{S}patio-\textbf{T}emporal \textbf{A}ggregation Across Views \textbf{T}rans\textbf{F}or\textbf{M}er), a multi-view 3D human pose estimation framework that predicts 3D poses from multi-view sequences of 2D poses (see Figure~\ref{fig:sta_tfm_pipeline}). This framework is specifically optimized for applications where capturing accurate joint angles and relative body geometry is the primary objective. For each view, STA-TFM extracts spatio-temporal embeddings using DSTformer \cite{zhu2023motionbert}, the feature extraction backbone of MotionBERT, originally designed for monocular pose estimation. These representations are then aggregated across views using a Fusion Pose Transformer (FPT) adapted from MPL \cite{ghasemzadeh2024mpl}. This design is motivated by the fact that monocular pose models can leverage abundant single-camera datasets, which are considerably easier to collect and annotate than synchronized multi-view captures. We show that their learned spatio-temporal representations can be effectively reused in a multi-view setting.

To the best of our knowledge, STA-TFM is the first method to leverage a frozen pretrained monocular feature extractor (DSTFormer) for multi-view 3D HPE. By decoupling per-view feature extraction from cross-view fusion, the proposed architecture simplifies training and allows optimization to focus on multi-view aggregation and 3D pose regression. Combined with the proposed fusion strategy, STA-TFM achieves improved motion consistency and 3D pose accuracy in flexible-layout environments without requiring explicit geometric priors or camera parameters during training or inference. Furthermore, the results suggest that monocular spatio-temporal representations learned by DSTFormer remain effective when transferred to multi-view 3D human pose estimation.

%% file: MDPI_Article_Template/Sections/2_related_works.tex
\label{sec:related_works}

Multi-view human pose estimation methods typically have higher accuracies as they effectively address the issues of depth ambiguity and occlusions prevalent with monocular HPE \cite{ghasemzadeh2024mpl, 
zheng2023deeplearningbasedhumanpose}. These models rely on multiple camera views, decreasing the chance of a fully occluded joint in all images. The information from additional views allows simple techniques, such as triangulation \cite{hartley2004multiple}, to be applied. Although triangulation is often the baseline for assessing multi-view setups, it is limited to cases where all keypoints are present and accurate \cite{ghasemzadeh2024mpl}. Additionally, accurate camera extrinsic and intrinsic parameters need to be known beforehand.

To address some of the limitations of triangulation, learning-based architectures have emerged. These methods can be broadly divided into camera-parameter methods, which require known camera intrinsics and extrinsic parameters, and camera-parameter-free methods, which do not.

\subsection{Camera-Parameter Methods}
Camera-parameter methods generally achieve higher accuracy than camera-parameter-free methods due to incorporating important prior knowledge into the network \cite{cai2024fusionformer, shuai2022adaptive}.
Epipolar Transformers \cite{he2020epipolar} exploit known camera parameters to identify corresponding epipolar lines across views, aggregating features from multiple cameras by attending to spatial locations along the epipolar line in one view using a joint query from another. Similarly, Transfusion \cite{ma2021transfusion} encodes 3D positional information into a transformer using an epipolar field, providing an efficient mechanism for encoding cross-view pixel correspondences. 

Volumetric methods, such as VoxelPose \cite{tu2020voxelpose}, project 2D joint heatmaps from all views into a shared 3D voxel space, using a Cuboid Proposal Network (CPN) to localize individuals and a Pose Regression Network (PRN) to estimate fine-grained 3D joint locations using 3D convolutional layers. SelfPose3D \cite{srivastav2024selfpose3d} extends the volumetric paradigm to a self-supervised setting, eliminating the need for 2D or 3D ground truth annotations by adopting a learning-by-projection paradigm, where the model learns 3D outputs by comparing projected 3D joint predictions against their 2D counterparts. While effective, volumetric approaches such as \cite{tu2020voxelpose, srivastav2024selfpose3d} incur significant computational cost \cite{choudhury2023tempo}. 

MVP \cite{zhang2021direct} addresses some of those inefficiencies by directly regressing multi-person 3D poses from multi-view images without relying on 2D pose priors, representing skeleton joints as learnable query embeddings that 
progressively attend to multi-view image features via a projective 
attention mechanism. MVGFormer \cite{liao2024multiple} takes a hybrid approach, combining a learning-free geometric module, which leverages triangulation to constrain the learning task, with a learnable appearance module for estimating 2D poses from images. 

TEMPO \cite{choudhury2023tempo} further extends calibrated methods to the spatio-temporal domain using a recurrent architecture to jointly learn spatial and temporal representations across views, achieving state-of-the-art results and demonstrating the importance of incorporating both time and multi-view cameras.

\subsection{Camera-Parameter-Free Methods}
Obtaining accurate camera parameters remains a significant hurdle in dynamic, uncontrolled settings \cite{shuai2022adaptive, gordon2022flex}, limiting the practical deployment of camera-parameter methods. Therefore, various methods attempt to remove the need for known camera extrinsics to obtain 3D pose from multi-view set ups.

Multi-view 3D Pose Lifter (MPL) \cite{ghasemzadeh2024mpl} demonstrated strong performance for multi-view fusion, harnessing two transformers, a spatial pose transformer and a fusion pose transformer (FPT). MPL effectively extracts meaningful spatial features from each view and fuses the embeddings to capture spatial features across views. 
Similarly, EFMK \cite{zhang2025efmk} combines Local-Global Pose Embedding with a Spatial-View Joint Transformer that integrates kinematic and geometric priors into attention computation. EFMK enforces bone length symmetry through a Bone-wise Alternating Optimization mechanism, and optionally refines predictions using reprojection-based multi-view aggregation when camera intrinsics are available.
However, both MPL and EFMK do not incorporate temporal information, which has been shown to improve accuracy and coherence of predicted 3D poses in both the monocular \cite{zhu2023motionbert} and multi-view setting \cite{choudhury2023tempo}.

MTF-Transformer \cite{shuai2022adaptive} addresses this by employing a Relative-Attention mechanism to measure implicit relationships between camera views using a Multi-view Fusing Transformer (MFT). This is followed by a Temporal Fusing Transformer (TFT) that aggregates features across the full sequence for final 3D pose prediction. However, this sequential spatial-then-temporal fusion restricts cross-domain communication between the two representations \cite{cai2024fusionformer}.

ESMformer \cite{zhang2025esmformer} also leverages relative attention and performs hierarchical fusion across spatial, temporal, cross-view, and cross-level pose features. Given multi-view 2D pose sequences, it extracts multi-level spatial features per-view using a relative-attention-based encoder, performs multi-view intra-level fusion across views and time, and then applies cross-level fusion before regressing the final 3D pose. While its error-aware self-supervised training reduces reliance on 3D annotations by using triangulation and reprojection losses, these geometric constraints still require known camera parameters during training.

Other methods such as \cite{cai2024fusionformer, gordon2022flex, zhang2024deep} also incorporate spatio-temporal features for multi-view 3D human pose estimation. FusionFormer \cite{cai2024fusionformer} encodes 2D HPE results into pose features, fuses them using a transformer encoder into a global spatio-temporal representation, and employs a transformer decoder that takes both global and view-specific features to estimate the 3D pose. FLEX \cite{gordon2022flex} takes multi-view 2D poses as input and uses a deep convolutional network to learn 3D rotations and bone lengths rather than joint locations directly, incorporating cross-view attention and temporal information. SGRAFormer \cite{zhang2024deep} employs a deep semantic graph transformer encoder to enrich spatial feature information by mining structural and skeletal edge knowledge of joints and their correlations. This is followed by a multi-view spatio-temporal fusion framework to mitigate joint depth uncertainty.

Recent advancements have introduced an intermediate category that internally estimates camera geometry, eliminating the need for explicitly provided external camera parameters during inference. EasyRet3D \cite{yin2025easyret3d} reconstructs human meshes independently from each view and uses Perspective-n-Point with RANSAC to initialize camera extrinsics from 3D-to-2D correspondences. Camera, body, and ground-plane parameters are then jointly refined. Qin et al. \cite{Qin2026Unconstrained} instead predict camera intrinsics and extrinsics directly from multi-view features using a transformer-based architecture and use the estimated parameters for differentiable triangulation. Additional algebraic and temporal constraints are incorporated to improve pose quality. While both methods eliminate the need for externally supplied camera calibration during inference, they still rely on a camera-estimation stage with sufficient cross-view correspondence. Furthermore, EasyRet3D relies on heavy sequence-level test-time optimization that limits real-time utility.

In contrast, STA-TFM directly regresses 3D poses from multi-view features without an intermediate camera parameter estimation step. STA-TFM takes multi-view 2D keypoints as input and feeds them to a frozen DSTformer \cite{zhu2023motionbert} to extract per-view spatio-temporal features. These embeddings are then passed to a fusion network (FPT) inspired by MPL \cite{ghasemzadeh2024mpl} to fuse cross-view features before a regression head produces the final 3D pose estimate. 

In contrast to existing methods like MTF-Transformer and FusionFormer that train their entire spatio-temporal and multi-view fusion components end-to-end, STA-TFM leverages a pre-trained, frozen DSTformer feature extractor from monocular HPE. By keeping the DSTformer frozen, we inherit strong temporal modeling capabilities learnt from training on large-scale monocular datasets while focusing optimization exclusively on the fusion network and regression head for cross-view aggregation.

%% file: MDPI_Article_Template/Sections/3_methodology.tex
\label{sec:methodology}

\begin{figure}[H]
%\isPreprints{\centering}{} % Only used for preprints
\centering
\includegraphics[width=\textwidth]{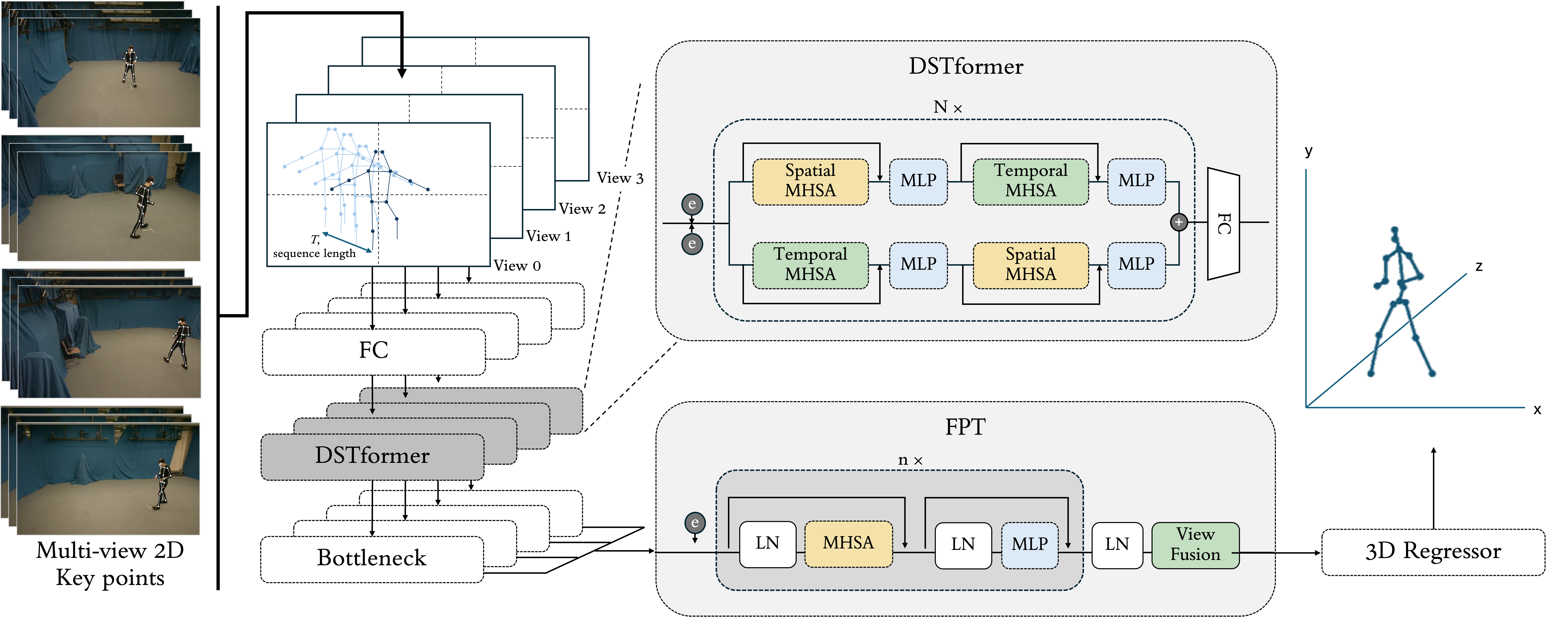}
\caption{STA-TFM architecture overview. Starting with multi-view 2D poses generated by an off-the-shelf pose estimator, the framework processes them through parallel DSTformer blocks to generate spatio-temporal embeddings for each view. These embeddings are compressed via bottleneck layers and concatenated before being passed to the Fusion Pose Transformer, which aggregates cross-view information using learnable positional embeddings. The final 3D pose is predicted through a lightweight regression head. Example shown is from the TotalCapture dataset \cite{trumble2017total}.\label{fig:architecture}}
\end{figure}

\subsection{Architecture}
\label{subsec:architecture}

The lifting architecture involves two key steps to transform 2D poses into 3D representations: DSTformer and a Fusion Pose Transformer (FPT). As shown in Figure \ref{fig:architecture}, the keypoints obtained from a generic 2D pose estimator (e.g. RTMPose \cite{jiang2023rtmpose}) are first passed to DSTformer to generate a motion representation, and a bottleneck layer is applied to compress the embedding before concatenating the embeddings from each view. The concatenated embedding is passed to a FPT to generate a fused embedding before the 3D pose is predicted by the final regression head.

DSTformer effectively captures the spatial and temporal information of the 2D skeleton within each view independently ~\cite{zhu2023motionbert}. With DSTformer, an input skeleton, $\text{x} \in \mathbb{R} ^{T \times J \times C_\text{in}}$, with sequence length $T$, $J$ joints, and input channel size, $C_\text{in}=3$, ($x$, $y$, $conf$), is first projected to a high dimensional feature space $\text{F}^{0} \in \mathbb{R} ^{T \times J \times C_\text{f}}$ and two learnable spatial and temporal positional encodings are added. Here, $C_\text{f}$ is the number of high dimensional features. DSTformer takes $\text{F}^{0}$ and passes it to the primary backbone.  This backbone includes sequential blocks consisting of a dual stream of alternating temporal and spatial multi-head self-attention (MHSA) with skip connections. Spatial MHSA attends to the joint relationships within the skeleton, while temporal MHSA models the relationships of single joints over time. The streams are subsequently fused using adaptive weighting. The DSTformer backbone of dual stream alternating MHSA is repeated $N$ times ($N=5$,  obtained through hyperparameter tuning) to get $\text{F}^{N} \in \mathbb{R} ^{T \times J \times C_\text{f}}$. A shared DSTformer backbone is applied independently to each camera view, producing one spatio-temporal embedding per view. These embeddings are compressed with a bottleneck layer before stacking the views $\text{E} \in \mathbb{R} ^{T \times J \times V \times C_\text{b}}$ for the FPT, where $V$ is the number of views (default=4), and $C_\text{b}$ is the bottleneck dimension.

The FPT architecture ~\cite{ghasemzadeh2024mpl} aggregates multiple per-view pose features into a single latent representation capturing structural and temporal cues across cameras. Given the stacked embeddings of DSTformer, the joints and channels are flattened to obtain a compatible input $\text{E} \in \mathbb{R} ^{T \times V \times (J*C_{b})}$, for the FPT. Learnable embeddings are added to the input to distinguish the cameras and provide positional context for each view. The fused tokens are then passed to a series of transformer blocks, each with MHSA that operates on the view dimension to learn inter-view relationships. 

The fusion transformer incorporates mask-aware attention to selectively attenuate contributions from missing or low-confidence camera inputs during cross-view information aggregation. By applying view masks to the attention weights $\alpha_{ij} \in \mathbb{R}^{V \times V}$ during cross-view aggregation, the model prevents invalid tokens from influencing attention computation. In combination with stable attention normalization, the model preserves numerical stability and performance under heterogeneous or incomplete multi-camera inputs.

Following the transformer blocks, the resulting features are normalized before applying a convolutional weighted mean across the view axis to obtain a unified embedding, $\text{E} \in \mathbb{R} ^{T \times (J*C_{b})}$. The embedding is reshaped, $\text{E} \in \mathbb{R} ^{T \times J \times C_{b}}$, before passing to a lightweight regression head to obtain the final output $\text{X}^\text{pred} \in \mathbb{R} ^{T \times J \times 3}$.

\subsection{Datasets}

The DHP19 ~\cite{calabrese2019dhp19}, HAA4D ~\cite{tseng2022haa4d}, BlendMimic3D \cite{lino20243d}, CMUPanoptic ~\cite{Joo_2017_TPAMI}, and TotalCapture ~\cite{trumble2017total} datasets are selected to represent human-centric motion across diverse environments. Table \ref{tab:datasets} outlines the different datasets and their specific use in this work. 

DHP19 and HAA4D are used to generate the training set for STA-TFM as described in Section \ref{subsec:synthetic_data}. DHP19 depicts standard human motion in controlled laboratory environments. It  consists of 17 subjects performing 33 motions recorded using 4 orthogonal cameras with Vicon ground truth for 13 joints. Conversely, HAA4D \cite{tseng2022haa4d} includes 3,390 samples and 300 action classes captured in uncontrolled environments. The videos are hand-labeled and processed using EvoSkeleton \cite{li2020cascaded} to obtain 17 3D joint positions, providing diverse and dynamic motion scenarios.

\begin{figure}[H]
%\isPreprints{\centering}{} % Only used for preprints
\centering
\includegraphics[width=\textwidth]{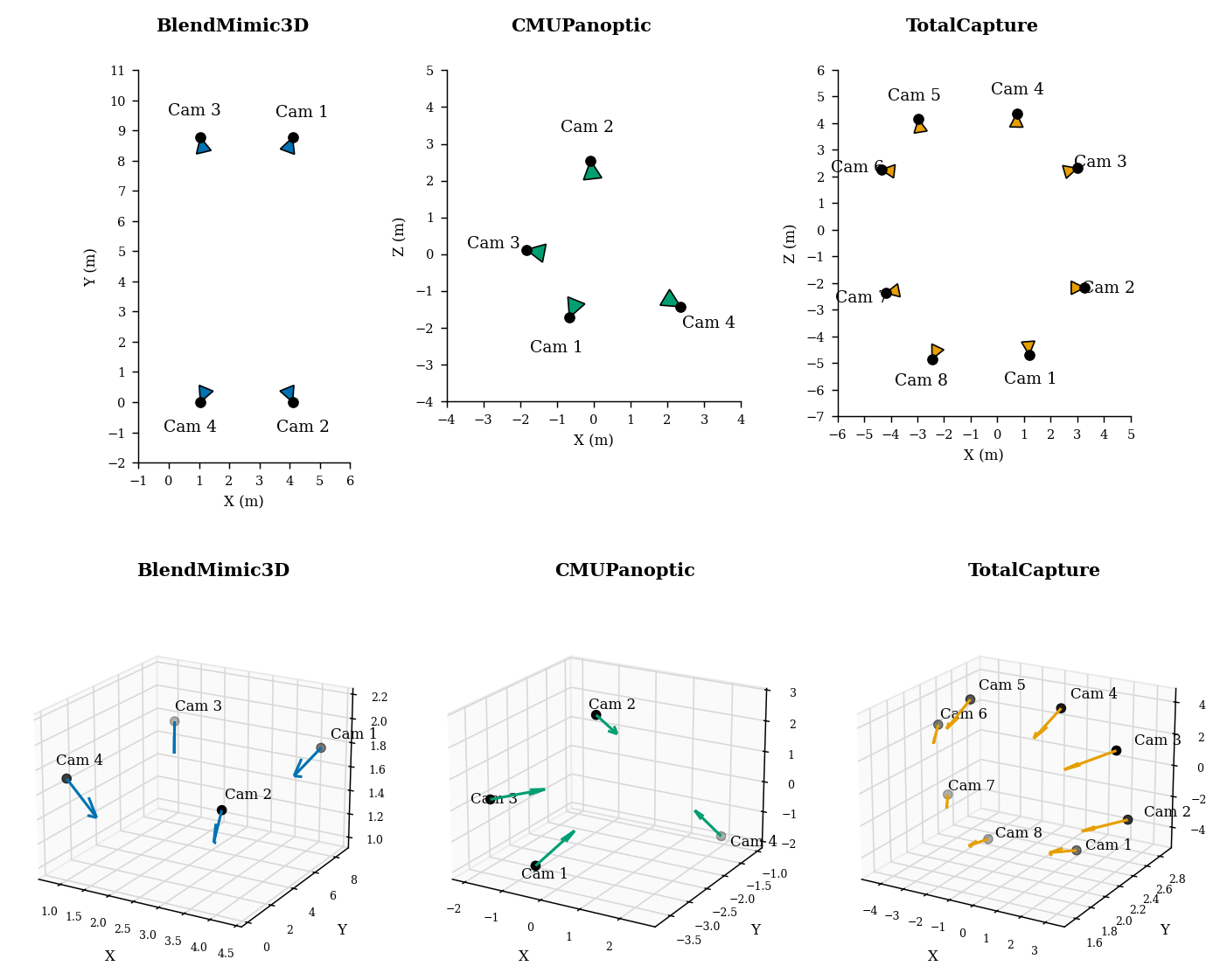}
\caption{Overhead (top) and 3D (bottom) camera placement layouts for Blendmimic3D, CMUPanoptic, and TotalCapture datasets. Triangles and arrows indicate camera position and viewing direction; axes are in metres.\label{fig:dataset-camera-layouts}}
\end{figure}

Additional datasets are also included to assess the peformance of STA-TFM and benchmark it against other methods. BlendMimic3D \cite{lino20243d} is a synthetic animated dataset generated using Blender \cite{Blender_2025}, consisting of three subjects performing various actions in different scenes. Each scenario is captured from a four-camera convergent setup, with evaluation in this work restricted to scenario S1. CMUPanoptic \cite{Joo_2017_TPAMI} is a multi-view dataset providing synchronized video from five cameras that capture multiple subjects in dense, multi-person scenarios. Finally, TotalCapture \cite{trumble2017total} offers synchronized multi-view video, IMU data, and ground truth 3D poses captured by a Vicon motion capture system. In total, this dataset consists of 1.9M frames spanning multiple subjects and activities, designed for 3D pose estimation evaluation. Figure \ref{fig:dataset-camera-layouts} illustrates the diverse camera configurations across these datasets.

While Human3.6M \cite{h36m_pami} is commonly used for benchmarking, licensing constraints prevented its inclusion. Instead, our diverse dataset selection enables a comprehensive assessment of generalization and performance across synthetic, controlled, and in-the-wild scenarios.

\begin{table}[H]
\caption{Summary of datasets used for training and evaluation. (\dag) indicates datasets transformed into synthetic multi-view scenarios for both training and evaluation, as described in Section \ref{subsec:synthetic_data}.\label{tab:datasets}}
%\isPreprints{\centering}{% This command is only used for ``preprints''.
	\begin{adjustwidth}{-\extralength}{0cm}
%} % If the paper is ``preprints'', please uncomment this parenthesis.
%\isPreprints{\begin{tabularx}{\textwidth}{CCC X}}{% This command is only used for ``preprints''.
		\begin{tabularx}{\fulllength}{C C C 2X}
%} % If the paper is ``preprints'', please uncomment this parenthesis.
			\toprule
			\textbf{Dataset} & \textbf{Type} & \textbf{Environment} & \textbf{Use}\\
			\midrule
			HAA4D * & Monocular & In-the-wild & Generating multi-view dataset + testing STA-TFM\\
			DHP19 * & Multi-view & Laboratory & Generating multi-view dataset + testing STA-TFM\\
			BlendMimic3D & Multi-view & Animated & Testing generalization to unseen camera setups\\
			CMUPanoptic & Multi-view & Laboratory & Benchmarking against camera-parameter methods\\
			TotalCapture & Multi-view & Laboratory & Benchmarking against camera-parameter-free methods\\
			\bottomrule
		\end{tabularx}
%		\isPreprints{}{% This command is only used for ``preprints''.
	\end{adjustwidth}
%} % If the paper is ``preprints'', please uncomment this parenthesis.
	\noindent{\footnotesize{* Datasets transformed into synthetic multi-view scenarios for both training and evaluation.}}
\end{table}

\subsection{Training Data Generation}
\label{subsec:synthetic_data}

\begin{figure}[H]
%\isPreprints{\centering}{} % Only used for preprints
\centering
\includegraphics[width=\textwidth]{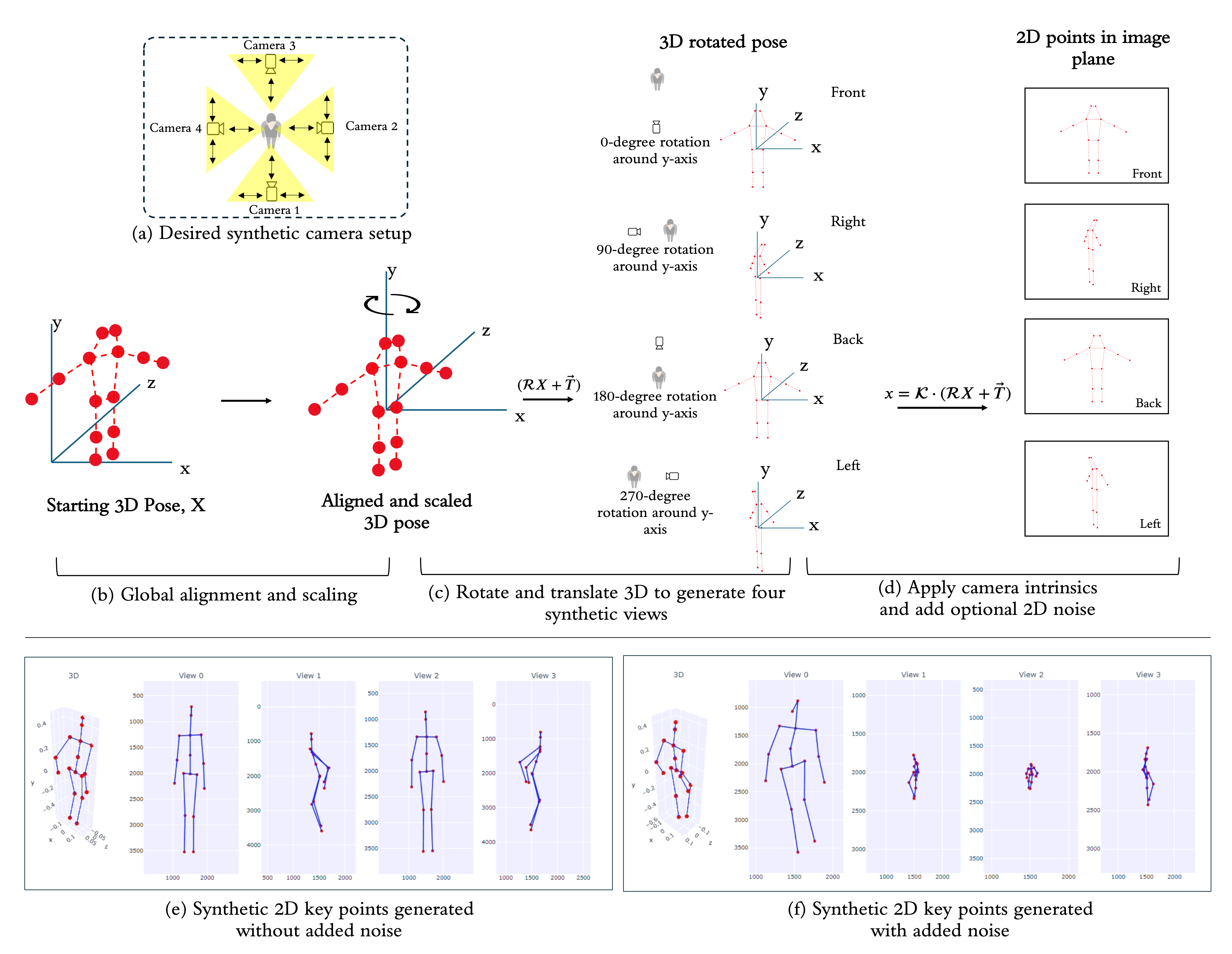}
\caption{Training data generation framework. (\textbf{a}) Camera placement with four cameras arranged radially around the subject. Yellow regions indicate approximate camera locations simulated by adding noise to camera extrinsics. (\textbf{b}) 3D poses from DHP19 and HAA4D datasets are globally aligned and scaled. (\textbf{c}) Aligned poses are transformed according to noisy camera extrinsics to simulate realistic setups with approximate camera placement. (\textbf{d}) 2D poses are obtained by applying perspective projection with camera intrinsics. (\textbf{e}) Clean generated 2D keypoints with zero noise. (\textbf{f}) Noisy generated 2D keypoints under realistic noise conditions.\label{fig:synthetic_data}}
\end{figure}

To address the challenges of collecting labeled multi-view human pose data, we use a data generation pipeline that converts any 3D pose collection into synthetic multi-view environments, as shown in Figure \ref{fig:synthetic_data}. 

% This approach enables any dataset containing 3D poses to serve as the foundation for multi-view training data. 
In this work, we utilize 3D poses from DHP19 and HAA4D as our base data.
Given potential skeletal topology differences across datasets, we first establish a consistent joint representation. Let $\mathcal{X} = {X_j \in \mathbb{R}^3 \mid j = 1,2,...,J}$ denote the set of 3D joint positions for a skeleton with $J$ joints. We interpolate specific anatomical landmarks to ensure consistency. The root joint position $X_{\text{root}}$ is defined as:

\begin{equation} 
            X_{\text{root}} = \frac{1}{2}(X_{\text{LHip}} + X_{\text{RHip}}),
\label{eq:interpolate_kp}
\end{equation}
\noindent where $X_{\text{LHip}}$ and $X_{\text{RHip}}$ are the locations of the left and right hip, respectively.

Processing involves three key steps: (1) global alignment, (2) scaling, and (3) perspective projection. 

\subsubsection{Global Alignment}
The 3D skeleton keypoints from the datasets, $X_j$, are first globally centered at the location of the root, $X_{\text{root}}$, to ensure the subject remains within the synthetic image space, as shown in Equation \eqref{eq:global_align}. Since multiple datasets are employed, this mapping aims to create a consistent coordinate system to reduce inter-dataset differences. 

\begin{equation} 
            X_j^{\text{aligned}} = X_j - X_{\text{root}}
\label{eq:global_align}
\end{equation}

%The skeletons are then locally normalized based on the maximum dimension of the skeleton for each frame, $t$, in the sequence, as defined in Equation \eqref{eq:local_norm}:

\subsubsection{Scaling}
Next, we apply scale normalization to standardize skeleton dimensions. For each frame $t$ in a sequence, we compute a scaling factor $s_t$ based on the maximum spatial extent across all coordinate axes, as defined in Equation \eqref{eq:local_norm}:

\begin{equation}
\begin{aligned}
    s_t &= \max_{c \in \{\text{x,y,z}\}} \left( \max_{j=1..J} \left( X_{j,t}^{(c)} \right) - \min_{j=1..J} \left( X_{j,t}^{(c)} \right) \right)\\
    X_{j,t}^{\text{scaled}} &= X_{j,t}^{\text{aligned}} \cdot \frac{1}{s_t}, \quad \forall j \in \{1,2,...,J\}
\end{aligned},
\label{eq:local_norm}
\end{equation}

\noindent where $X_{j,t}$ is the 3D key point for joint $j$ at time $t$. $J$ is the total number of joints. $c$ is the coordinate axis (x, y, or z).

\subsubsection{Perspective Projection}
The final step involves generating 2D keypoints for multiple views to simulate the outputs of a 2D pose estimator applied across cameras. An imaging environment is simulated, with four cameras arranged radially around the origin. 
% at approximately 0\textdegree, 90\textdegree, 180\textdegree, and 270\textdegree. 
To simulate different setups, we add random noise to the extrinsic parameters,  $E$ (Equation \eqref{eq:extrinsic}) defining the camera position and orientation, specifically to the rotation matrix $\mathcal{R} \in \mathbb{R}^{3 \times 3}$ and to the translation vector $\Vec{T} \in \mathbb{R}^{3 \times 1}$ of the camera with respect to the global coordinate system. 

\begin{equation} 
E = [\mathcal{R} | \Vec{T}]
\label{eq:extrinsic} 
\end{equation}

The rotation matrix, $\mathcal{R} $ is composed as shown in Equation \eqref{eq:rotation_comp}:

\begin{equation} 
    \mathcal{R} = R_z(\gamma + \delta_\gamma) R_y(\beta + \delta_\beta) R_x(\alpha + \delta_\alpha) ,
\label{eq:rotation_comp} 
\end{equation}

\noindent where the matrices $R_z(.)$, $R_y(.)$, and $R_x(.)$ are the elementary rotation matrices around the $z$-, $y$-, and $x$-axis, respectively, and $\delta_\gamma, \delta_\beta, \delta_\alpha \sim \mathcal{U}[-\epsilon_\text{angle}, \epsilon_\text{angle}]$ are the corresponding camera angle noise terms.

The translation vector, $\Vec{T}$ is composed as shown in Equation \eqref{eq:dist_vector}:

\begin{equation} 
\begin{aligned}
\Vec{T} &= -\mathcal{R}c\\
c &= -\mathcal{R}^T \begin{bmatrix} 0 \\ 0 \\ d + \delta_d \end{bmatrix}, 
\end{aligned}
\label{eq:dist_vector} 
\end{equation}

\noindent where $d$ is the desired camera distance and $\delta_d \sim \mathcal{U}[-\epsilon_\text{dist}, \epsilon_\text{dist}]$ is the corresponding distance noise term.

The intrinsic parameters, represented by matrix $\mathcal{K}$, are based on a standard iPhone camera, with random noise sampled uniformly from the interval $[-\epsilon_\text{int}, \epsilon_\text{int}]$ added to the focal length ($f_x$, $f_y$) and principal point ($c_x$, $c_y$). Equation \ref{eq:intrinsics} describes the intrinsics matrix:

% \begin{equation}
% \mathcal{K} = \begin{bmatrix}
% f_x & 0 & c_x \\
% 0 & f_y & c_y \\
% 0 & 0 & 1
% \end{bmatrix}.
% \label{eq:intrinsics}
% \end{equation}

\begin{equation}
\mathcal{K} = \begin{bmatrix}
f_x + \delta_{f} & 0 & c_x + \delta_{c} \\
0 & f_y + \delta_{f} & c_y + \delta_{c} \\
0 & 0 & 1
\end{bmatrix}.
\label{eq:intrinsics}
\end{equation}

\noindent where $\delta_{f}, \delta_{c} \sim \mathcal{U}[-\epsilon_\text{int}, \epsilon_\text{int}]$ are the corresponding intrinsics noise terms.

Intrinsic and extrinsic parameters are constant in each view and regenerated for each sequence. Perspective projection is applied to obtain simulated 2D keypoints, $x$, for each view. As shown in Equation \eqref{eq:perspective_proj}, this projection accounts for the camera intrinsic and extrinsic properties to simulate realistic 2D poses from any sample 3D pose:
\begin{equation}
x = \mathcal{K} \cdot (\mathcal{R}X + \Vec{T}).
\label{eq:perspective_proj}
\end{equation}

To mimic noisy predictions from a 2D HPE model, Gaussian noise proportional to body height is added to each keypoint in view $v$:

\begin{equation}
x_{j,v}^{\text{noisy}} = x_{j,v} + \eta_{j,v}, \quad \eta_{j,v} \sim \mathcal{N}(0, \sigma_{j,v}^2),
\label{eq:keypoint_noise}
\end{equation}

\noindent where $\sigma_{j,v} = \epsilon_{\text{joint},v} \cdot h_v$ is the noise standard deviation, $h_v$ is the body height in pixels for view $v$, and $\epsilon_{\text{joint},v}$ is the noise scale parameter that varies across views.

The combined noise in camera intrinsic and extrinsic parameters simulates realistic deployments where cameras are radially distributed, positioned at varying distances, and subjected to different pitch and roll angles. This training strategy improves STA-TFM's generalization to diverse camera placements without requiring camera parameters to be specified. Additionally, since the subject moves freely throughout each sequence, there is no fixed notion of a `front' or `back' camera, preventing the network from learning view-specific anatomical shortcuts. In practice, arranging cameras in an ordered sequence around the subject remains important for optimal performance.

\subsection{Model Training}
\label{subsec:training}
The synthetic dataset described in Section \ref{subsec:synthetic_data} is divided into 70\% for training and validation and 30\% for testing, where the training subset is further split with a 20\% validation ratio. The normalized samples are windowed to a length of 36 frames with a stride of 12, ensuring overlap within a sequence. 

During training, we implement two loss metrics: a standard spatial loss and a composite spatial-temporal loss.  Mean Per Joint Position Error (MPJPE) is the primary loss metric and measures the average Euclidean distance between the predicted and ground truth joints. The Mean Per Joint Velocity Error (MPJVE) is a supplemental loss metric, and accounts for temporal differences in joints across frames. The composite loss, $\mathcal{L_\text{composite}}$, as shown in Equation ~\eqref{eq:comp_loss} is used to account for both the spatial and temporal features of the model during training, where $\lambda \in [0,1]$ controls the strength of the MPJVE component. Through hyperparameter tuning, we found $\lambda = 1$ yields the best results.
\begin{equation}
\begin{aligned}
    \mathcal{L_\text{composite}} &=  \mathcal{L}_\text{MPJPE} + \lambda\mathcal{L}_\text{MPJVE} \\
     \mathcal{L}_\text{MPJPE} &= \frac{1}{T}\frac{1}{J} \sum_{t=0}^{T} \sum_{j=0}^{J} \left\lVert X_{j,t}^{\text{pred}} - X_{j,t}^{\text{scaled}} \right\rVert_2 \\
     \mathcal{L}_\text{MPJVE} &= \frac{1}{T-1}\frac{1}{J} \sum_{t=0}^{T-1} \sum_{j=0}^{J} \left\lVert V_{j,t}^{\text{pred}} - V_{j,t}^{\text{scaled}} \right\rVert_2 
\end{aligned}
\label{eq:comp_loss}
\end{equation}

Here, the velocity $V_{j,t}$ is as defined in Equation \eqref{eq:velocity}, $T$ is the video sequence length, $J$ is the number of joints supported in the dataset.
\begin{equation}
   V_{j,t} = X_{j,t+1} - X_{j,t}
\label{eq:velocity}
\end{equation}

We train the model on a single NVIDIA Tesla V100-SXM2-16GB GPU with 4 physical (8 logical) CPU cores. Training employs automated mixed precision and gradient clipping ($||\Delta\leq1||$). The AdamW\cite{AdamW} optimizer is used with an initial learning rate of $1 \times 10^{-4}$ and cosine annealing for weight decay. Each run uses a batch size of 16 and is trained for 5 epochs. Training was limited to 5 epochs to ensure fair comparison with MPL under identical conditions, despite training curves indicating potential for further improvement. STA-TFM's architecture contains 51M total parameters, with 35M trainable parameters. The complete training process takes approximately 2.5 hours on 73,590 training samples with 4-camera, 36-frame sequences. Inference on the same hardware achieves 37.73 ms per instance (36-frame sequence, 4 cameras).

As there is a temporal embedding mismatch between DSTformer (window length = 243) and STA-TFM (window length = 36), linear interpolation is required to reduce the temporal embedding length. The DSTformer pretrained backbone is initialized from pretrained weights from \cite{zhu2023motionbert} and could optionally be fine-tuned, while the Fusion Pose Transformer and 3D regression head weights are learned. 

To better generalize to occlusions, camera flicker, and view dropout, stochastic masking is applied during training. A dropout rate of 0.25 is applied to each view to simulate environments with missing cameras. If all cameras are dropped, a single camera is randomly re-initialized to ensure at least one view remains. To simulate camera flicker, a per-frame dropout rate of 0.01 is also applied.

\subsection{Evaluation Metrics}
\label{subsec:eval_metrics}
To evaluate the predicted 3D keypoints in comparison to ground truth (GT) data, we use commonly reported metrics for human pose estimation outlined in~\cite{pavllo20193d,guo2025survey, tu2020voxelpose}: Mean Per Joint Position Error (MPJPE), Percentage of Correct Keypoints (PCK), Mean Per Joint Velocity Error (MPJVE). Additionally, Average Precision ($\text{AP}_K$) is computed for comparison with other camera-parameter and camera-parameter-free methods.

The MPJPE calculates the average Euclidean distance between predicted and ground-truth joint locations across all frames. Smaller MPJPE values indicate better performance. 
Formally, it is defined as:

\begin{equation}
    \text{MPJPE} = \frac{1}{T} \frac{1}{J} \sum_{t=1}^{T} \sum_{j=1}^J \|\hat{X}_{j,t} - X_{j,t}\|_2
\end{equation}
where $T$ is the number of frames, $J$ is the number of joints, $\hat{X}_{j,t}$ is the predicted position of joint $j$ at time $t$, and $X_{j,t}$ is the corresponding ground truth position.

PCK assesses accuracy by checking if predicted keypoints are within a certain distance threshold from the real joints, with the threshold set to 20\% of the torso length. This is calculated as:
\begin{equation}
    \text{PCK} = \frac{1}{T} \frac{1}{J} \sum_{t=1}^{T} \sum_{j=1}^J \mathbbm{1}\left(\|\hat{X}_{j,t} - X_{j,t}\|_2 \leq 0.2 \cdot l_{\text{torso}}\right) \times 100\%
\end{equation}
where $l_{\text{torso}}$ is the torso length and $\mathbbm{1}(\cdot)$ is the indicator function that returns 1 if the condition is true and 0 otherwise.

To evaluate temporal consistency and the smoothness of predictions over time, the MPJVE measures the average error of the first derivative of 3D pose sequences. It is defined as:
\begin{equation}
    \text{MPJVE} = \frac{1}{T-1} \frac{1}{J} \sum_{t=1}^{T-1} \sum_{j=1}^J \|\hat{V}_{j,t} - V_{j,t}\|_2
\end{equation}
where $V_{j,t} = X_{j,t+1} - X_{j,t}$ represents the velocity of joint $j$ at time $t$, and $\hat{V}_{j,t}$ and $V_{j,t}$ are the predicted and ground truth velocities, respectively.

Lastly, $\text{AP}_K$ \cite{tu2020voxelpose} measures 3D pose estimation accuracy by computing precision-recall curves at a distance threshold set to $K$ mm. Each predicted pose is matched to the closest ground truth by MPJPE. A prediction is a true positive if its MPJPE < $K$ mm and the matched ground truth has not been previously detected, otherwise it is a false positive. AP is computed as the area under the precision-recall curve, evaluated at thresholds $K \in {25, 50, 100, 150}$ mm, with the mean reported as mAP.

STA-TFM is trained to predict zero-centered 3D skeletons centered at the root. Since the predicted skeletons are normalized and root-centered, they exist in a different coordinate frame than the ground truth, making direct comparison without alignment uninformative. Procrustes alignment is therefore applied to account for variations in translation, rotation, and uniform scaling between the predicted and ground truth skeletons. Importantly, Procrustes alignment does not modify the relative positions between joints or the overall skeleton shape, preserving inter-joint relationships and ensuring that the alignment step does not artificially inflate performance. This yields the primary evaluation metrics: P-MPJPE, P-PCK, and P-MPJVE. This approach is appropriate for applications such as movement assessment, joint angle estimation, and range of motion analysis, where accurate inter-joint relationships and body geometry are the primary objectives rather than absolute location in space.

%% file: MDPI_Article_Template/Sections/4_experiments.tex
\label{sec:experiments}

We evaluate STA-TFM across five datasets to assess its performance under different conditions. First, we evaluate the 2D-to-3D lifting capabilities of STA-TFM, examining the impact of temporal incorporation and composite loss function on datasets derived from DHP19 and HAA4D. Second, we assess generalizability to novel datasets using BlendMimic3D, which was not seen during training. We then benchmark STA-TFM against both camera-parameter and camera-parameter-free methods on two widely used benchmarks, CMUPanoptic and TotalCapture. To simulate real-world constraints, we further analyze the model's performance under varying camera counts and 2D keypoint noise. We then examine the impact of temporal sequence length, concluding with a comparative analysis against leading monocular baselines. Finally, we present a qualitative assessment of STA-TFM to demonstrate its multi-view 3D pose reconstruction capabilities

\input{MDPI_Article_Template/Sections/4_1_temporal_analysis}  
\input{MDPI_Article_Template/Sections/4_2_impact_loss}

\input{MDPI_Article_Template/Sections/4_4_unseen_datasets}

\input{MDPI_Article_Template/Sections/4_5_cam_param_free_methods}

\input{MDPI_Article_Template/Sections/4_6_cam_param_methods}  
\input{MDPI_Article_Template/Sections/4_7_impact_of_num_cameras}

\input{MDPI_Article_Template/Sections/4_8_impact_of_noise}

\input{MDPI_Article_Template/Sections/4_3_impact_of_sequence_length}

\input{MDPI_Article_Template/Sections/4_9_monocular_analysis}
\input{MDPI_Article_Template/Sections/4_10_qualitative_analysis}

%% file: MDPI_Article_Template/Sections/4_1_temporal_analysis.tex
\subsection{Impact of Temporal Modeling}
\label{subsection:temporal_analysis}
To quantify the performance gains of incorporating temporal information into pose embeddings, STA-TFM is compared against vanilla MPL ~\cite{ghasemzadeh2024mpl}, which lacks temporal information. %, and linear triangulation \cite{hartley2004multiple}, which estimates 3D keypoints from multi-view 2D keypoints. 
Table \ref{tab:comparison_to_prior} presents the performance comparison between these methods. To ensure a fair comparison, both methods are trained on the same dataset generated from HAA4D and DHP19, for the same number of epochs. For all experiments, predictions utilize four 2D skeletons sequences captured at approximately 90-degree intervals with varying distances from the subject. Each 2D skeleton is normalized and zero-centered around the root before processing. Since MPL lacks temporal integration, relying solely on spatial features, STA-TFM's improvement over the vanilla model is expected. Therefore, to establish a fair baseline, a naive Savitzky-Golay temporal filter is applied post-hoc to MPL's predictions.

Table \ref{tab:comparison_to_prior} shows that STA-TFM significantly outperforms vanilla MPL on both DHP19 and HAA4D datasets. Even when a Savitzky-Golay temporal filter is applied post-hoc to MPL's predictions, improvements remain marginal and inconsistent across datasets, suggesting that a learned approach to temporal modeling is more effective than naive post-processing.

On DHP19, STA-TFM achieves 50.9\% and 49.5\% reductions in P-MPJPE and P-MPJVE, respectively, compared to MPL, with similar improvements of 6.7\% and 7.7\% on HAA4D. These gains in spatial accuracy and temporal coherence are directly attributed to incorporating temporal information, which provides an inherent denoising effect by considering multiple time points.
% \hl{far more effectively than naive post-processing.}

The performance variation between datasets can be attributed to several key factors. First, the training data distribution contains a natural imbalance, with DHP19 comprising approximately 80\% of the training frames. Second, the datasets employ different splitting strategies: DHP19 uses subject-based splits that maintain action diversity during training, while HAA4D employs action-based splits that require generalization to unseen motion patterns. HAA4D also presents additional challenges through its more diverse and dynamic actions and unconstrained environments. 

Despite these challenging conditions in HAA4D, the temporal modeling in STA-TFM still provides improvements over MPL, demonstrating the value of integrating temporal information even for novel actions.

\begin{table}[H]
\caption{Performance comparison on DHP19, HAA4D, and BlendMimic3D datasets. All models are trained on a dataset generated from a union of HAA4D and DHP19 datasets. Lower $\downarrow$ indicates better performance; higher $\uparrow$ indicates better accuracy. Best results are bolded. ($\dag$) Indicates novel dataset, unseen during training. $w$ indicates the window size for the Savitzky-Golay filter that yields the best results.
\label{tab:comparison_to_prior}}
%\isPreprints{\centering}{% This command is only used for ``preprints''.
	\begin{adjustwidth}{-\extralength}{0cm}
%} % If the paper is ``preprints'', please uncomment this parenthesis.
%\isPreprints{\begin{tabularx}{\textwidth}{CCCCCC}}{% This command is only used for ``preprints''.
		\begin{tabularx}{\fulllength}{XXXXXC}
%} % If the paper is ``preprints'', please uncomment this parenthesis.
			\toprule
			\textbf{Dataset} & \textbf{Model} & \textbf{Loss Function} & \textbf{P-MPJPE (mm) $\downarrow$} & \textbf{P-PCK (\%) $\uparrow$} & \textbf{P-MPJVE (mm/frame) $\downarrow$}\\
			\midrule
\multirow[m]{4}{*}{DHP19} & STA-TFM & Composite Loss & \textbf{17.56} & \textbf{99.96} & 1.66\\
			  	                   & STA-TFM & MPJPE Loss     & 18.71          & 99.93          & 2.82\\
			             	      & MPL     & Default        & 35.78          & 99.49          & 3.29\\
                                   & MPL+SG ($w$=11)& Default        & 35.72          & 99.49          & \textbf{1.28}\\
                   \midrule
\multirow[m]{4}{*}{HAA4D} & STA-TFM & Composite Loss & \textbf{34.77} & \textbf{98.70} & \textbf{8.55}\\
			  	                  & STA-TFM & MPJPE Loss     & 36.48          & 98.42          & 8.73\\
			             	     & MPL     & Default        & 37.27          & 98.44          & 9.26\\
                                   & MPL+SG ($w$=5) & Default        & 37.88          & 98.23          & 10.63\\
                   \midrule
\multirow[m]{3}{*}{BlendMimic3D $\dag$} & STA-TFM & Composite Loss & \textbf{61.97} & \textbf{85.92} & \textbf{2.00}\\
			  	                  & MPL     & Default        & 212.25         & 10.02          & 2.99\\
                                   & MPL+SG ($w$=5)& Default        & 212.25         & 10.02          & 3.00\\
			\bottomrule
		\end{tabularx}
%		\isPreprints{}{% This command is only used for ``preprints''.
	\end{adjustwidth}
%} % If the paper is ``preprints'', please uncomment this parenthesis.
	\noindent{\footnotesize{($\dag$) Indicates novel dataset, unseen during training.}}
\end{table}

%% file: MDPI_Article_Template/Sections/4_2_impact_loss.tex
\subsection{Impact of Loss Function}
MPJPE is the primary loss function typically used in 3D HPE model training. However, this loss function does not penalize 3D HPE predictions with low temporal consistency (jittery predictions). Therefore, STA-TFM improves temporal consistency by utilizing a composite loss, $\mathcal{L}_{\text{composite}}$, as described in Section~\ref{subsec:training}. To assess the impact of the composite loss, two versions of STA-TFM are trained: one with a standard loss based solely on MPJPE and another with $\mathcal{L}_{\text{composite}}$. Both models are trained on DHP19 + HAA4D, and results are reported on each dataset separately in Table~\ref{tab:comparison_to_prior}.

The results in Table~\ref{tab:comparison_to_prior} demonstrate a clear improvement when using $\mathcal{L}_{\text{composite}}$ on both DHP19 and HAA4D datasets. Specifically, STA-TFM achieves 41.0\% and 2.3\% reductions in P-MPJVE on DHP19 and HAA4D, respectively. An improvement in P-MPJPE is also observed, with reductions of 6.2\% and 5.0\% on DHP19 and HAA4D, respectively. This spatial accuracy improvement can be attributed to $\mathcal{L}_{\text{composite}}$ encouraging temporally consistent poses, which reduces frame-to-frame jitter and consequently improves per-frame predictions. The varying impact of $\mathcal{L}_{\text{composite}}$ on P-MPJVE between datasets reflects their inherent differences, as described in Section~\ref{subsection:temporal_analysis}.

%% file: MDPI_Article_Template/Sections/4_4_unseen_datasets.tex
\subsection{Generalization to Unseen Datasets}

% \subsubsection{{3D Pose Accuracy}}
% \label{subsubsec:pose_acc}
To assess generalizability to unseen datasets, STA-TFM and MPL are trained on DHP19 and HAA4D datasets and evaluated on BlendMimic3D, which features new motion patterns and unfamiliar camera placements and configurations. As shown in Table~\ref{tab:comparison_to_prior}, STA-TFM substantially outperforms MPL across all metrics, achieving 70.8\% and 41.6\% reductions in P-MPJPE and P-MPJVE, respectively. These improvements stem from STA-TFM's ability to use  pretrained DSTformer backbone, which provides transferable spatio-temporal representations that enhance generalization to unseen data. Since DSTFormer is adapted from monocular human pose estimation, it benefits from training on monocular datasets that are more abundant than multi-view datasets, contributing to improved generalization across diverse scenarios.

%% file: MDPI_Article_Template/Sections/4_5_cam_param_free_methods.tex
\subsection{Performance Against Camera-Parameter-Free Methods} % 

To evaluate STA-TFM against existing camera-parameter-free methods, we use the TotalCapture dataset following the training and testing split of \cite{shuai2022adaptive}. The evaluation protocol tests generalization across four scenarios defined by combinations of seen and unseen cameras and subjects. In all scenarios, the test samples are novel and were not included in the training set. Specifically: (1) seen cameras and seen subjects, where both the camera setup and subjects appeared in the training set but are performing different actions; (2) seen cameras and unseen subjects, where the camera setup was seen during training but the subjects are novel; (3) unseen cameras and seen subjects, where the subjects appeared in training but the camera configuration is novel; and (4) unseen cameras and unseen subjects, where both the camera setup and subjects are novel to the model.

While various comparable camera-parameter-free methods report results on TotalCapture \cite{shuai2022adaptive, cai2024fusionformer, ghasemzadeh2024mpl, zhang2024deep, zhang2025efmk}, many do not release their code bases, preventing direct comparison. We therefore limit our comparison to methods with publicly accessible code bases. Among these, none provide TotalCapture-specific model weights or evaluation code, requiring us to retrain all baselines on identical dataset splits using the hyperparameters specified in their original papers. The training set consists of 2D keypoints obtained from RTMPose \cite{jiang2023rtmpose} paired with their corresponding 3D ground truth skeletons, introducing realistic 2D keypoint predictor noise into the training and evaluation. 
Results are reported in Table \ref{tab:total_capture_comparison}. P-MPJPE is reported as the primary evaluation metric for all methods, ensuring that differences in alignment protocols are accounted for to provide a balanced comparison. Under this standard, STA-TFM achieves highly competitive results when looking at the mean of both seen and unseen subjects and camera setups, achieving 15.2\%, 35.7\%, 40.8\%, 50.2\%, and 55.22\% reductions in P-MPJPE compared to MPL \cite{ghasemzadeh2024mpl}, SGraFormer \cite{zhang2024deep}, MTF-Transformer \cite{shuai2022adaptive}, EFMK \cite{zhang2025efmk}, and ESMFormer \cite{zhang2025esmformer}, respectively. As expected, all methods experience a performance drop when evaluated on unseen subjects or cameras.

\begin{table}[H]
\caption{Performance comparison on the TotalCapture dataset. Values show P-MPJPE in millimeters. All models use identical training data with 2D keypoints from RTMPose and hyperparameters from the original papers. ($\dag$) denotes modified hyperparameters for improved performance. Action types: W (walking), FS (free style), A (acting). Best results are \textbf{bolded}.\label{tab:total_capture_comparison}}
%\isPreprints{\centering}{% This command is only used for ``preprints''.
	\begin{adjustwidth}{-\extralength}{0cm}
%} % If the paper is ``preprints'', please uncomment this parenthesis.
%\isPreprints{\begin{tabularx}{\textwidth}{C CCC CCC C}}{% This command is only used for ``preprints''.
		\begin{tabularx}{\fulllength}{2C XXXXXXX}
%} % If the paper is ``preprints'', please uncomment this parenthesis.
			\toprule
			\multirow{2}{*}{\textbf{Method}} & \multicolumn{3}{c}{\textbf{Seen Subjects (S1, S2, S3)}} & \multicolumn{3}{c}{\textbf{Unseen Subjects (S4, S5)}} & \multirow{2}{*}{\textbf{Mean}}\\
			\cmidrule(lr){2-4} \cmidrule(lr){5-7}
			 & \textbf{W2} & \textbf{FS3} & \textbf{A3} & \textbf{W2} & \textbf{FS3} & \textbf{A3} & \\
			\midrule
			\multicolumn{8}{c}{\textbf{Seen Cameras (1,3,5,7)}}\\
			\midrule
			EFMK \cite{zhang2025efmk} & 48.94 & 89.24 & 55.61 & 55.03 & 112.62 & 66.45 & 71.31\\
			MTF-Transformer \cite{shuai2022adaptive} & 32.99 & 80.59 & 57.50 & 56.89 & 105.58 & 76.26 & 68.30\\
			MTF-Transformer \cite{shuai2022adaptive} ($\dag$) & 30.43 & 74.97 & 56.20 & 50.53 & 102.37 & 73.00 & 64.50\\
			ESMFormer \cite{zhang2025esmformer} & 44.65 & 72.49 & 55.11 & 53.36 & 93.16 & 60.95 & 63.29\\
			SGraFormer \cite{zhang2024deep} & 15.46 & 41.55 & 24.98 & 27.93 & 73.99 & 40.61 & 37.42\\
			MPL \cite{ghasemzadeh2024mpl} & 18.19 & 42.18 & 26.21 & 26.76 & 68.80 & 39.43 & 36.93\\
			STA-TFM & \textbf{13.15} & \textbf{40.64} & \textbf{22.65} & \textbf{25.98} & \textbf{64.17} & \textbf{37.65} & \textbf{34.04}\\
			\midrule
			\multicolumn{8}{c}{\textbf{Unseen Cameras (2,4,6,8)}}\\
			\midrule
			EFMK \cite{zhang2025efmk} & 104.90 & 139.13 & 104.90 & 99.01 & 138.10 & 99.37 & 99.24\\
			MTF-Transformer \cite{shuai2022adaptive} & 69.78 & 101.49 & 75.67 & 81.30 & 112.45 & 88.38 & 88.17\\
			MTF-Transformer \cite{shuai2022adaptive} ($\dag$) & 59.12 & 94.90 & 66.50 & 63.76 & 108.47 & 81.66 & 79.06\\
            ESMFormer \cite{zhang2025esmformer} & 103.31 & 142.76 & 110.87 & 115.32 & 157.65 & 128.39 & 126.38\\
			SGraFormer \cite{zhang2024deep} & 82.54 & 101.87 & 91.29 & 89.86 & 111.19 & 91.32 & 94.67\\
			MPL \cite{ghasemzadeh2024mpl} & 51.01 & 73.78 & 56.91 & 56.27 & 85.47 & 56.14 & 63.26\\
			STA-TFM & \textbf{32.21} & \textbf{61.18} & \textbf{40.87} & \textbf{42.04} & \textbf{77.90} & \textbf{51.10} & \textbf{50.90}\\
			\bottomrule
		\end{tabularx}
%		\isPreprints{}{% This command is only used for ``preprints''.
	\end{adjustwidth}
%} % If the paper is ``preprints'', please uncomment this parenthesis.
\label{tab:total_capture_comparison}
\end{table}

%% file: MDPI_Article_Template/Sections/4_6_cam_param_methods.tex
\subsection{Performance Against Camera-Parameter Methods} 

CMUPanoptic is a popular benchmark for evaluating camera-parameter methods. While these baseline models provide a valuable performance ceiling across the broader literature, they are not directly comparable to camera-parameter-free approaches. This is because they operate in a different problem space by relying on known camera 
parameters as geometric priors, allowing them to operate in absolute world space. On the other hand, STA-TFM functions as a camera-parameter-free method, focusing on capturing accurate body geometry in normalized space. Consequently, the evaluation metrics differ: calibrated models report absolute MPJPE, while normalized outputs require Procrustes aligned MPJPE (P-MPJPE).

To contextualize the performance of STA-TFM against these models, both STA-TFM and MPL are trained on CMUPanoptic following the training and testing split of VoxelPose \cite{tu2020voxelpose}. We include MPL, the most architecturally comparable camera-parameter-free method to STA-TFM, as a reference point for comparison. Training is performed using 2D keypoints from RTMPose \cite{jiang2023rtmpose} paired with their corresponding 3D ground truth skeletons. 

CMUPanoptic contains multi-person scenarios, but STA-TFM like all other comparable camera-parameter-free methods \cite{ghasemzadeh2024mpl, zhang2024deep, shuai2022adaptive, zhang2025efmk} supports only single-person inference. To handle this, we implemented a pre-processing step where 2D keypoints from all camera views were cross-associated to identify and group keypoints belonging to the same person, isolating individual subjects before passing them to the models for inference.

Table \ref{tab:cmu_panoptic} presents STA-TFM and MPL against several camera-parameter methods. Despite the advantage of using Procrustes alignment, as expected, camera-parameter methods achieve higher accuracy than camera-parameter-free methods due to incorporating geometric priors \cite{cai2024fusionformer, shuai2022adaptive}. Nevertheless, STA-TFM outperforms MPL ($\downarrow$ 16.5\% P-MPJPE, $\uparrow$ 1.12\% PCK, $\downarrow$ 41.65\% MPJVE). To further benchmark STA-TFM against traditional non-learning camera-parameter methods, the framework is also evaluated against a classical linear triangulation baseline, with full methodological details and comparative results provided in \ref{app:triangulation}.

\begin{table}[H]
%\small % Change table font size
\caption{Comparison of camera-parameter and camera-parameter-free methods on the CMUPanoptic dataset. MPL and STA-TFM use RTMPose 2D keypoints as input. ($\dag$) denotes model performance using noiseless ground truth 2D keypoints as input. $\text{AP}_K$ is computed from precision-recall curves at distance threshold $K$ mm, where a prediction is a true positive if its MPJPE $< K$ mm. For camera-parameter-free methods, reported metrics are computed after applying Procrustes alignment between the predicted and ground truth poses. \label{tab:cmu_panoptic}}
%\isPreprints{\centering}{} % Only used for preprints
\begin{adjustwidth}{-\extralength}{0cm}
    \begin{tabularx}{\fulllength}{CCCCCCCCC}
    \toprule
    \textbf{Method} & \textbf{PCK} & \textbf{MPJVE} & \textbf{AP25} & \textbf{AP50} & \textbf{AP100} & \textbf{AP150} & \textbf{Recall @500} & \textbf{MPJPE (mm)}\\
    \midrule
    \multicolumn{9}{c}{\textbf{Multi-view methods with camera parameters}}\\
    \cmidrule(lr){1-9}
    VoxelPose~\cite{tu2020voxelpose}          & - & - & 83.60  & 98.30  & 99.80  & 99.90  & 98.80  & 17.70\\
    MVP~\cite{zhang2021direct}                & - & - & 92.30  & 96.60  & 97.50  & 97.70  & 98.20  & 15.80\\
    TEMPO~\cite{choudhury2023tempo}            & - & - & 89.01 & 99.08 & 99.76 & 99.93 & -     & 14.68\\
    MVGFormer~\cite{liao2024multiple}         & - & - & 92.30  & -     & -     & -     & -     & 16.00\\
    SelfPose3D~\cite{srivastav2024selfpose3d} & - & - & 55.10  & 96.40  & 98.50  & 99.00  & 99.60  & 24.50\\
    \midrule
    \multicolumn{9}{c}{\textbf{Multi-view methods without camera parameters}}\\
    \cmidrule(lr){1-9}
    MPL~\cite{ghasemzadeh2024mpl}             & 95.78 & 5.81 & 5.90  & 63.96 & 98.73 & 99.79 & 100   & 36.15\\
    STA-TFM                                   & 96.86 & 3.39 & 19.70  & 76.70  & 99.40  & 99.90  & 99.90  & 30.17\\
   \mbox{STA-TFM~($\dag$)}                    & 96.76 & 3.08 & 25.00  & 68.00  & 99.20  & 99.70  & 99.90  & 28.83\\
    \bottomrule
    \end{tabularx}
\end{adjustwidth}
\noindent{\footnotesize{($\dag$) Model performance using noiseless ground truth 2D keypoints as input.}}
\end{table}

\subsection{Impact of Fine-Tuning DSTFormer} 
\label{subsec:ablation_dstformer}
To assess whether fine-tuning DSTFormer yields additional performance gains for STA-TFM, we compare frozen and unfrozen variants trained for 50 epochs on 
CMUPanoptic with identical hyperparameters. Table~\ref{tab:frozen_unfrozen} shows that unfrozen training yields negligible improvements (MPJPE: 
$30.17 \rightarrow 30.26$~mm with RTMPose keypoints, $28.83 \rightarrow 28.29$~mm with GT keypoints), demonstrating that monocular spatio-temporal features transfer effectively to multi-view 3D HPE without the need for task-specific fine-tuning.

\begin{table}[H]
\caption{Ablation study comparing frozen and unfrozen DSTFormer variants of STA-TFM 
on CMUPanoptic. Both variants are trained for 50 epochs with identical hyperparameters, 
initialized from pretrained monocular DSTFormer weights. For all methods, reported 
metrics are computed after applying Procrustes alignment between the predicted and 
ground truth poses. \label{tab:frozen_unfrozen}}
\begin{adjustwidth}{-\extralength}{0cm}
    \begin{tabularx}{\fulllength}{CCCCCCCCC}
    \toprule
    \textbf{DSTFormer} & \textbf{\begin{tabular}[c]{@{}c@{}}2D\\Keypoints\end{tabular}} & \textbf{PCK} & \textbf{MPJVE} & \textbf{AP25} & \textbf{AP50} & \textbf{AP100} & \textbf{AP150} & \textbf{MPJPE (mm)}\\
    \midrule
    Frozen   & RTMPose & 96.86 & 3.39 & 19.70 & 76.70 & 99.40 & 99.90 & 30.17\\
    Frozen   & GT      & 96.76 & 3.08 & 25.00 & 68.00 & 99.20 & 99.70 & 28.83\\
    Unfrozen & RTMPose & 96.79 & 3.35 & 20.00 & 77.30 & 99.40 & 99.90 & 30.26\\
    Unfrozen & GT      & 96.88 & 2.92 & 32.40 & 68.40 & 99.40 & 99.90 & 28.29\\
    \bottomrule
    \end{tabularx}
\end{adjustwidth}
\end{table}

%% file: MDPI_Article_Template/Sections/4_7_impact_of_num_cameras.tex
%%%%%%%%%%%%%%%%%%%%%%%%%%%%%%%%%%%%%%%%%%%%%%%%%%%%%%%%%%%
\subsection{Impact of Number of Cameras in Multi-View Methods}
\begin{figure}[H]
%\isPreprints{\centering}{} % Only used for preprints
\centering
\includegraphics[width=\textwidth]{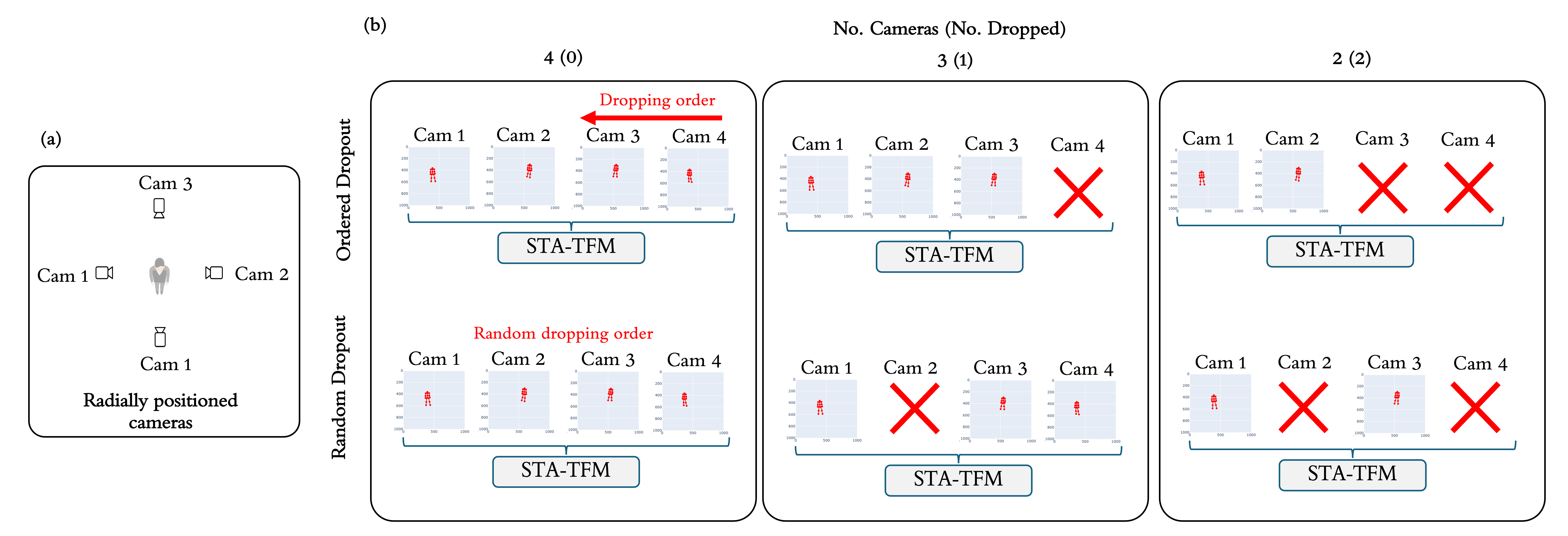}
\caption{(a) Ideal camera placement. (b) Camera dropout strategies. Starting from four cameras positioned at 0\degree, 90\degree, 180\degree, and 270\degree, we evaluate two removal patterns: (Top) Consecutive dropout removes the last $m$ cameras in order (e.g., 270\degree, then 180\degree). (Bottom) Random dropout removes $m$ cameras at random positions (e.g., 90\degree, then 270\degree). Red cross indicates cameras that are dropped.\label{fig:cam_dropout}}
\end{figure}

STA-TFM requires a minimum of two distinct camera views to predict 3D keypoints from a given set of 2D keypoints. Ideally, these cameras should be radially positioned for optimal performance (see Figure \ref{fig:cam_dropout}a). To assess the impact of the number of views on 3D keypoint prediction accuracy, Table \ref{tab:num_cameras_analyis} reports STA-TFM performance relative to the number of cameras. Starting from a 4-camera setup, we simulate reduced coverage by removing cameras using two strategies: consecutive dropout, where the last $m$ consecutive cameras ($\sim$90$^\circ$ apart) are removed, and random dropout, where $m$ cameras are randomly dropped regardless of their position. Figure \ref{fig:cam_dropout}b demonstrates both dropout strategies.

Table \ref{tab:num_cameras_analyis} shows that performance decreases with fewer cameras, with the four-camera setup providing the lowest error rates. The results also reveal that camera order significantly impacts accuracy: consecutive camera dropout yields better performance than random dropout. This can be attributed to the learnable positional embeddings in the fusion transformer, which encode the spatial relationships between views. When cameras are dropped consecutively, the remaining views maintain their expected positional context. However, random dropout disrupts these learned spatial relationships, leading to degraded performance.

\begin{table}[H]
\caption{Performance comparison on DHP19 and HAA4D datasets for different numbers of cameras and camera dropout strategies. Lower $\downarrow$ indicates better performance; higher $\uparrow$ indicates better accuracy. Best results are bolded.\label{tab:num_cameras_analyis}}
%\isPreprints{\centering}{% This command is only used for ``preprints''.
	\begin{adjustwidth}{-\extralength}{0cm}
%} % If the paper is ``preprints'', please uncomment this parenthesis.
%\isPreprints{\begin{tabularx}{\textwidth}{CCCCCC}}{% This command is only used for ``preprints''.
		\begin{tabularx}{\fulllength}{CCCCCC}
%} % If the paper is ``preprints'', please uncomment this parenthesis.
			\toprule
			\textbf{Dataset} & \textbf{Dropout Strategy} & \textbf{No. Cameras (No. Dropped)} & \textbf{P-MPJPE (mm) $\downarrow$} & \textbf{P-PCK (\%) $\uparrow$} & \textbf{P-MPJVE (mm/frame) $\downarrow$}\\
			\midrule
\multirow[m]{5}{*}{DHP19} & No dropout & 4 (0) & \textbf{17.56} & \textbf{99.96} & \textbf{1.66}\\
\cmidrule(lr){2-6}
			  	                   & \multirow[m]{2}{*}{Ordered} & 3 (1) & 37.51 & 97.83 & 1.94\\
			  	                   &                             & 2 (2) & 74.45 & 86.16 & 2.35\\
\cmidrule(lr){2-6}
			  	                   & \multirow[m]{2}{*}{Random}  & 3 (1) & 43.87 & 96.83 & 2.13\\
			  	                   &                             & 2 (2) & 120.66 & 69.88 & 5.30\\
                   \midrule
\multirow[m]{5}{*}{HAA4D} & No dropout & 4 (0) & \textbf{34.77} & \textbf{98.70} & \textbf{8.55}\\
\cmidrule(lr){2-6}
			  	                  & \multirow[m]{2}{*}{Ordered} & 3 (1) & 68.26 & 86.62 & 10.66\\
			  	                  &                             & 2 (2) & 109.44 & 60.73 & 13.27\\
\cmidrule(lr){2-6}
			  	                  & \multirow[m]{2}{*}{Random}  & 3 (1) & 72.71 & 83.98 & 10.80\\
			  	                  &                             & 2 (2) & 155.77 & 46.44 & 19.30\\
			\bottomrule
		\end{tabularx}
%		\isPreprints{}{% This command is only used for ``preprints''.
	\end{adjustwidth}
%} % If the paper is ``preprints'', please uncomment this parenthesis.
\end{table}

%% file: MDPI_Article_Template/Sections/4_8_impact_of_noise.tex
%%%%%%%%%%%%%%%%%%%%%%%%%%%%%%%%%%%%%%%%%%%%%%%%%%%%%%%%%%%
\subsection{Impact of 2D Pose Estimation Noise on 3D Prediction}

\begin{figure}[H]
    \centering
    \includegraphics[width=0.8\textwidth]{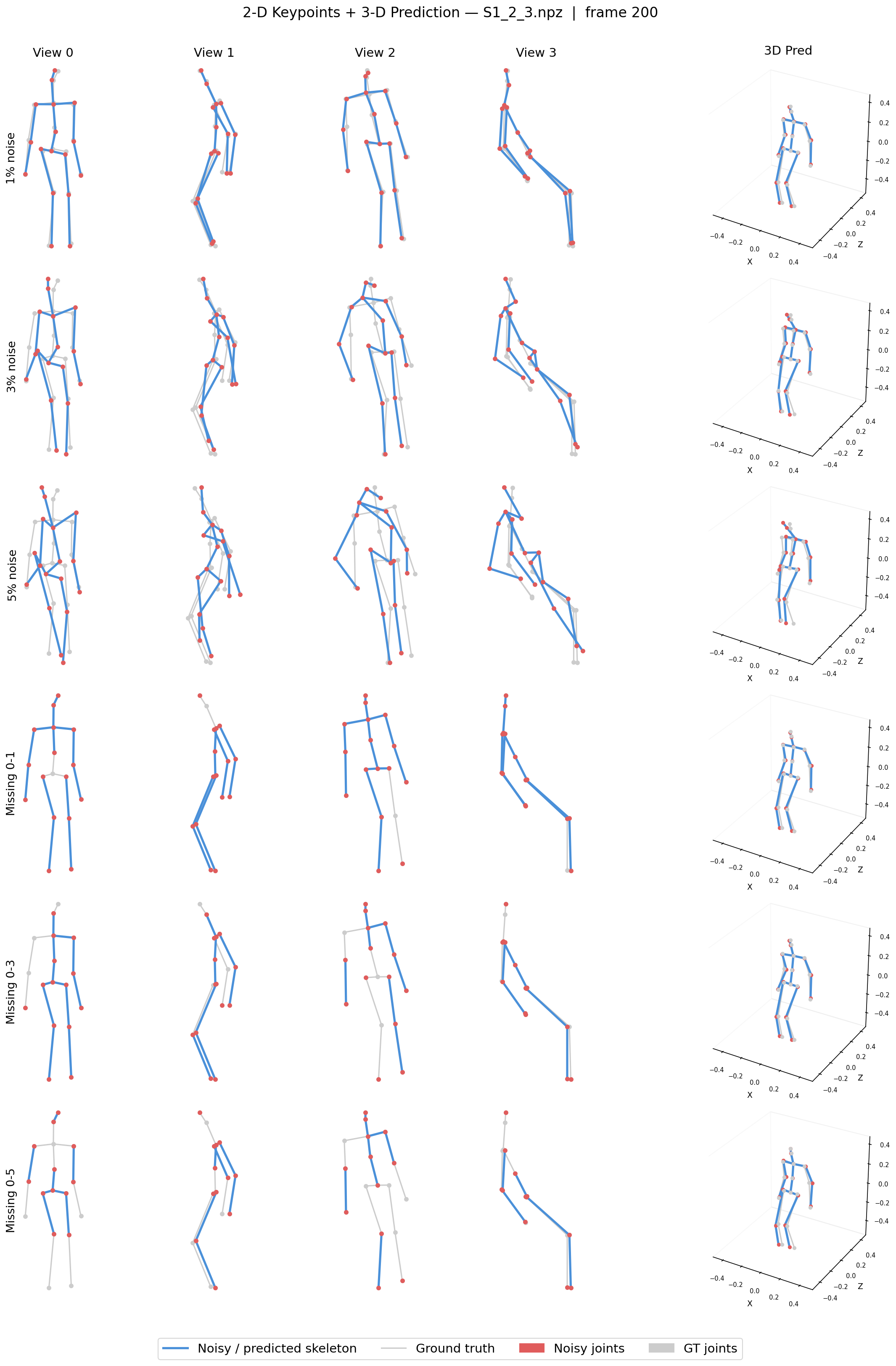}
    \caption{Effect of input corruption on 2D keypoint observations and 3D pose predictions. Rows show increasing noise levels: Gaussian perturbation at 1\%, 3\%, and 5\%, and keypoint removal of 0--1, 0--3, and 0--5 joints. Columns show the four camera views. The rightmost column shows the predicted 3D pose. Gray underlays indicate ground truth. Results shown for DHP19 sequence S1\_2\_3.}
    \label{fig:noise1}
\end{figure}

\begin{figure}[H]
    \centering
    \includegraphics[width=0.8\textwidth]{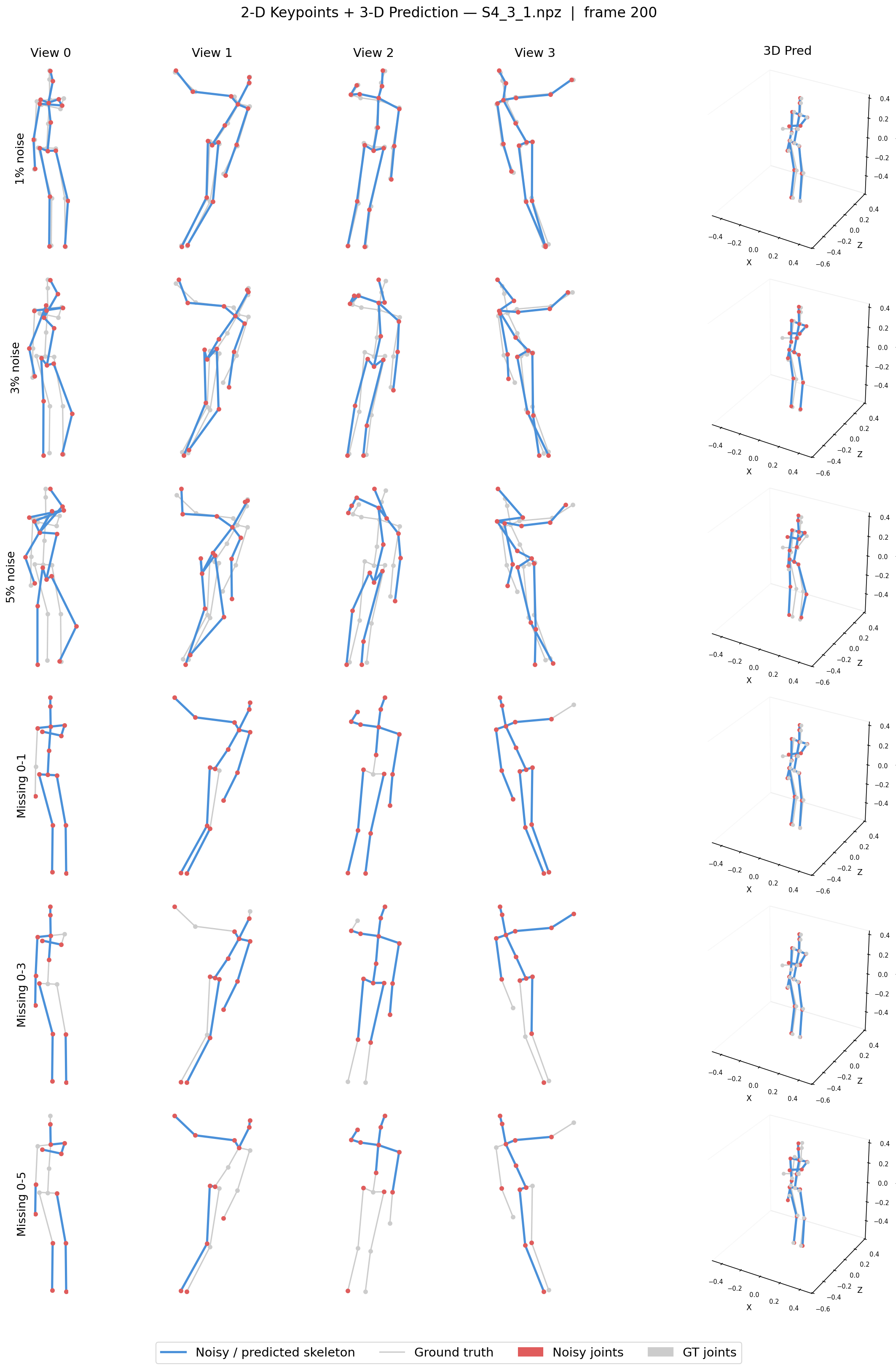}
    \caption{Effect of input corruption on 2D keypoint observations and 3D pose predictions. Rows show increasing noise levels: Gaussian perturbation at 1\%, 3\%, and 5\%, and keypoint removal of 0--1, 0--3, and 0--5 joints. Columns show the four camera views; the rightmost column shows the predicted 3D pose. Gray underlays indicate ground truth. Results shown for DHP19 sequence S4\_3\_1.}
    \label{fig:noise2}
\end{figure}

To investigate individual impacts of Gaussian noise and keypoint dropout, both were systematically injected into the DHP19 testing dataset. Table \ref{tab:2d_noise} presents a performance analysis across varying levels of additive Gaussian noise and keypoint dropout, simulating the effects of a noisy 2D keypoint detector and occluded joints. Noise level is expressed as a percentage of subject height, where Gaussian noise with a standard deviation of $\sigma = \text{height} \times p$ is added to each keypoint, with $p$ denoting the noise percentage. Keypoint dropout is simulated by randomly removing between $[n_\text{min}, n_\text{max}]$ keypoints per view per frame, where $(n_\text{min}, n_\text{max}) \in {(0,1), (0,3), (0,5)}$. Figures  \ref{fig:noise1}, \ref{fig:noise2} illustrate the noise levels experimented with alongside the corresponding predicted and ground truth 3D poses.

The model demonstrates reasonable tolerance to keypoint dropout, maintaining stable performance when up to five keypoints are randomly occluded per frame (P-MPJPE increases from 17.56$\rightarrow$23.83 mm), which covers typical occlusion scenarios in real-world capture.
For Gaussian noise, the model shows minimal degradation at 1\% of subject height (P-MPJPE: 18.28 mm), a noise level representative of modern 2D pose detectors like RTMPose. At 5\% noise, performance degrades substantially (P-MPJPE: 54.88 mm, P-MPJVE: 52.18 mm/frame), indicating that the model is sensitive to the larger positional uncertainties characteristic of low-quality 2D detections. This behavior is expected for camera-parameter-free methods, which lack the geometric constraints available to calibrated approaches and must rely more heavily on input quality.

\begin{table}[H]
%\small % Change table font size
\caption{Robustness analysis on the DHP19 dataset evaluating the effect of additive Gaussian noise and keypoint dropout on P-MPJPE, P-PCK, and P-MPJVE. Bold text indicates better performance.\label{tab:2d_noise}}
%\isPreprints{\centering}{} % Only used for preprints
\begin{tabularx}{\textwidth}{CCCCC}
\toprule
\textbf{\# Missing KP} & \textbf{\% Noise} & \textbf{P-MPJPE (mm) $\downarrow$} & \textbf{P-PCK (\%) $\uparrow$} & \textbf{P-MPJVE (mm/frame) $\downarrow$}\\
\midrule
0     & 0 & \textbf{17.56 $\pm$ 4.99} & \textbf{99.96 $\pm$ 0.22} & \textbf{1.66 $\pm$ 0.46}\\
\midrule
0     & 1 & 18.28 $\pm$ 4.92          & 99.96 $\pm$ 0.21          & 3.63 $\pm$ 0.98\\
0     & 3 & 30.76 $\pm$ 8.14          & 99.79 $\pm$ 0.57          & 22.24 $\pm$ 8.68\\
0     & 5 & 54.88 $\pm$ 10.96         & 95.96 $\pm$ 4.22          & 52.18 $\pm$ 13.64\\
\midrule
0--1  & 0 & 18.26 $\pm$ 4.81          & 99.95 $\pm$ 0.23          & 4.03 $\pm$ 0.79\\
0--3  & 0 & 20.42 $\pm$ 4.58          & 99.66 $\pm$ 0.37          & 6.88 $\pm$ 1.41\\
0--5  & 0 & 23.83 $\pm$ 4.61          & 97.77 $\pm$ 0.69          & 9.14 $\pm$ 1.87\\
\bottomrule
\end{tabularx}

\noindent{\footnotesize{Bold text indicates better performance.}}
\end{table}

%% file: MDPI_Article_Template/Sections/4_3_impact_of_sequence_length.tex
\subsection{Impact of Sequence Length}

As DSTFormer \cite{zhu2023motionbert} was originally pretrained using 243-frame windows, an ablation was performed to evaluate the effects of temporal window length when applied to datasets with variable sequence lengths. HAA4D contains short clips with an average length of 64 frames, while DHP19 contains substantially longer recordings, averaging 2110 frames. To accommodate shorter sequences, STA-TFM was trained with reduced window lengths, 81, 54, and 36 frames, while window strides remained a third of the sequence length. Table \ref{tab:seq_len} shows that DHP19 performance remains almost similar across window lengths, with a decrease in P-MPJPE ($\downarrow 3.78 \text{mm}$) but an increase in P-MPJVE ($\uparrow 0.09\text{mm/frame}$) when reducing the window length from 81 frames to 36 frames. This indicates that the model can effectively extract motion cues from temporal segments of multiple lengths. 

The primary performance benefit was seen with the short HAA4D videos as shown in Table \ref{tab:seq_len}. When windowed at 36 frames, the P-MPJPE ($\downarrow 11.97\text{mm}$) and P-MPJVE ($\downarrow 2.5\text{mm/frame}$) decreased significantly in comparison to a temporal length of 81. This improvement arises as shorter windows better match HAA4D’s intrinsic clip duration and reduce the influence of padding. These findings demonstrate that compact temporal windows can effectively capture motion in brief videos without compromising performance in long-duration videos.

\begin{table}[H]
%\small % Change table font size
\caption{Comparison of P-MPJPE, P-PCK, and P-MPJVE across sequence lengths on DHP19 and HAA4D. Bold indicates better performance.\label{tab:seq_len}}
%\isPreprints{\centering}{} % Only used for preprints
\begin{tabularx}{\textwidth}{CCCCC}
\toprule
\textbf{Dataset} & \textbf{Seq. Length} & \textbf{P-MPJPE (mm) $\downarrow$} & \textbf{P-PCK (\%) $\uparrow$} & \textbf{P-MPJVE (mm/frame) $\downarrow$}\\
\midrule
\multirow[m]{3}{*}{DHP19} & 81 & 21.34 $\pm$ 7.13                  & 99.80 $\pm$ 0.60                  & \textbf{1.57 $\pm$ 0.47}\\
                          & 54 & 22.61 $\pm$ 5.87                  & 99.90 $\pm$ 0.50                  & 1.61 $\pm$ 0.43\\
                          & 36 & \textbf{17.56 $\pm$ 4.99}         & \textbf{99.96 $\pm$ 0.22}         & 1.66 $\pm$ 0.46\\
\midrule
\multirow[m]{3}{*}{HAA4D} & 81 & 46.74 $\pm$ 7.13                  & 95.40 $\pm$ 8.40                  & 11.05 $\pm$ 11.50\\
                          & 54 & 43.47 $\pm$ 13.71                 & 97.00 $\pm$ 6.30                  & 11.09 $\pm$ 11.69\\
                          & 36 & \textbf{34.77 $\pm$ 11.45}        & \textbf{98.70 $\pm$ 4.21}         & \textbf{8.55 $\pm$ 8.23}\\
\bottomrule
\end{tabularx}

\noindent{\footnotesize{Bold indicates better performance.}}
\end{table}

%% file: MDPI_Article_Template/Sections/4_9_monocular_analysis.tex
\subsection{Multi-View vs. Single-View Analysis}
\label{supplementary:Monocular Methods}

To contextualize STA-TFM performance relative to monocular approaches, we compare against popular single-camera 3D HPE models: MotionBERT \cite{zhu2023motionbert}, RTMW3D \cite{jiang2024rtmw}, VideoPose3D \cite{pavllo20193d}, and BlazePose (MediaPipe) \cite{bazarevsky2020blazepose}. Multi-view methods inherently benefit from reduced depth ambiguity and improved occlusion handling compared to monocular approaches. We quantify this advantage in this analysis. These models are selected because they incorporate temporal information across multiple frames to predict 3D poses, ensuring a fair comparison. Since STA-TFM utilizes the DSTformer block from MotionBERT to generate spatially and temporally aware 2D pose embeddings, the comparison with MotionBERT is particularly relevant, as any performance gains can be directly attributed to multi-view fusion via the FPT block. Comparisons with other models are provided for completeness. To ensure fairness, STA-TFM is provided with 2D keypoints computed by running RTMPose \cite{jiang2023rtmpose} on videos from the four camera views in BlendMimic3D.

Table \ref{tab:multi_vs_single_dhp19} shows that STA-TFM achieves substantially better performance compared to the best monocular model (VideoPose3D): P-MPJPE ($\downarrow 11.6\%$), P-PCK ($+ 44.72\%$), and P-MPJVE ($\downarrow 57.1\%$). Similar improvements are observed compared to MotionBERT: P-MPJPE ($\downarrow 20.9\%$), P-PCK ($+47.12\%$), and P-MPJVE ($\downarrow 61.4\%$). As expected, this performance advantage decreases with fewer cameras, as shown in the 3-camera and 2-camera configurations in Table \ref{tab:multi_vs_single_dhp19}. Notably, when constrained to a two-camera configuration, STA-TFM does not outperform the leading monocular model in terms of P-MPJPE; however, it consistently outperforms monocular models in P-PCK and P-MPJVE. This suggests that even limited cross-view aggregation provides a regularizing effect on temporal jitter, enabling smoother and more anatomically consistent 3D motions than single-camera architectures can produce. These results demonstrate that multi-view information provides more accurate 3D poses compared to monocular approaches, which suffer from depth ambiguity and occlusions.

\begin{table}[H]
%\small % Change table font size
\caption{Performance comparison on BlendMimic3D for multi-view and monocular 3D HPE models. Lower $\downarrow$ is better; higher $\uparrow$ is better. Best results are bolded.\label{tab:multi_vs_single_dhp19}}
%\isPreprints{\centering}{} % Only used for preprints
\begin{tabularx}{\textwidth}{CCCCC}
\toprule
\textbf{No. Cameras} & \textbf{Model} & \textbf{P-MPJPE (mm) $\downarrow$} & \textbf{P-PCK (\%) $\uparrow$} & \textbf{P-MPJVE (mm/frame) $\downarrow$}\\
\midrule
4 & STA-TFM & \textbf{94.17 $\pm$ 17.38} & \textbf{70.07 $\pm$ 10.11} & \textbf{4.77 $\pm$ 2.46}\\
3 & STA-TFM & 98.39 $\pm$ 17.23 & 67.61 $\pm$ 10.25 & 4.86 $\pm$ 2.41\\
2 & STA-TFM & 108.02 $\pm$ 13.86 & 60.39 $\pm$ 9.06 & 4.45 $\pm$ 2.00\\
\midrule
1 & MotionBert \cite{zhu2023motionbert} & 119.04 $\pm$ 39.25 & 22.95 $\pm$ 22.38 & 12.84 $\pm$ 5.92\\
1 & RTMW3D \cite{jiang2024rtmw} & 125.81 $\pm$ 34.15 & 20.79 $\pm$ 20.05 & 32.24 $\pm$ 18.89\\
1 & VideoPose3D \cite{pavllo20193d} & \textbf{106.49 $\pm$ 30.68} & 25.35 $\pm$ 23.86 & 11.49 $\pm$ 5.29\\
1 & BlazePose \cite{bazarevsky2020blazepose} & 216.80 $\pm$ 27.56 & 6.49 $\pm$ 6.08 & 16.85 $\pm$ 8.63\\
1 & MotionAGFormer \cite{mehraban2024motionagformer} & 116.80 $\pm$ 28.44 & \textbf{38.59 $\pm$ 9.80} & \textbf{7.90 $\pm$ 4.57}\\
\bottomrule
\end{tabularx}
\end{table}

%% file: MDPI_Article_Template/Sections/4_10_qualitative_analysis.tex
\subsection{Qualitative Assessment}

Qualitative results of STA-TFM under challenging conditions are presented in Figure~\ref{fig:qualitative_assessment}. The associated full video sequences are provided in the supplementary material showing multi-view 3D pose reconstruction for subjects S1, S2, and S3 performing acting and freestyle sequences on the TotalCapture dataset.

Notably, when subjects move out of view and the 2D detector hallucinates non-human artifacts as keypoints, STA-TFM maintains reasonable 3D pose reconstruction (Figure~\ref{fig:qualitative_assessment}(a)). Furthermore, the model demonstrates tolerance to occluded joints in single views and successfully reconstructs poses from remaining cameras when one or more views contain missing keypoints (Figure~\ref{fig:qualitative_assessment}(b)). This behaviour is attributed to the integration of keypoint confidence scores into the DSTFormer feature embeddings, which down-weights unreliable detections. Additionally, the model's predictions are less sensitive to scale and translation, as training on root-centered and height-normalized skeletons removes these dependencies.

\begin{figure}[H]
%\isPreprints{\centering}{} % Only used for preprints
\centering
\includegraphics[width=\textwidth]{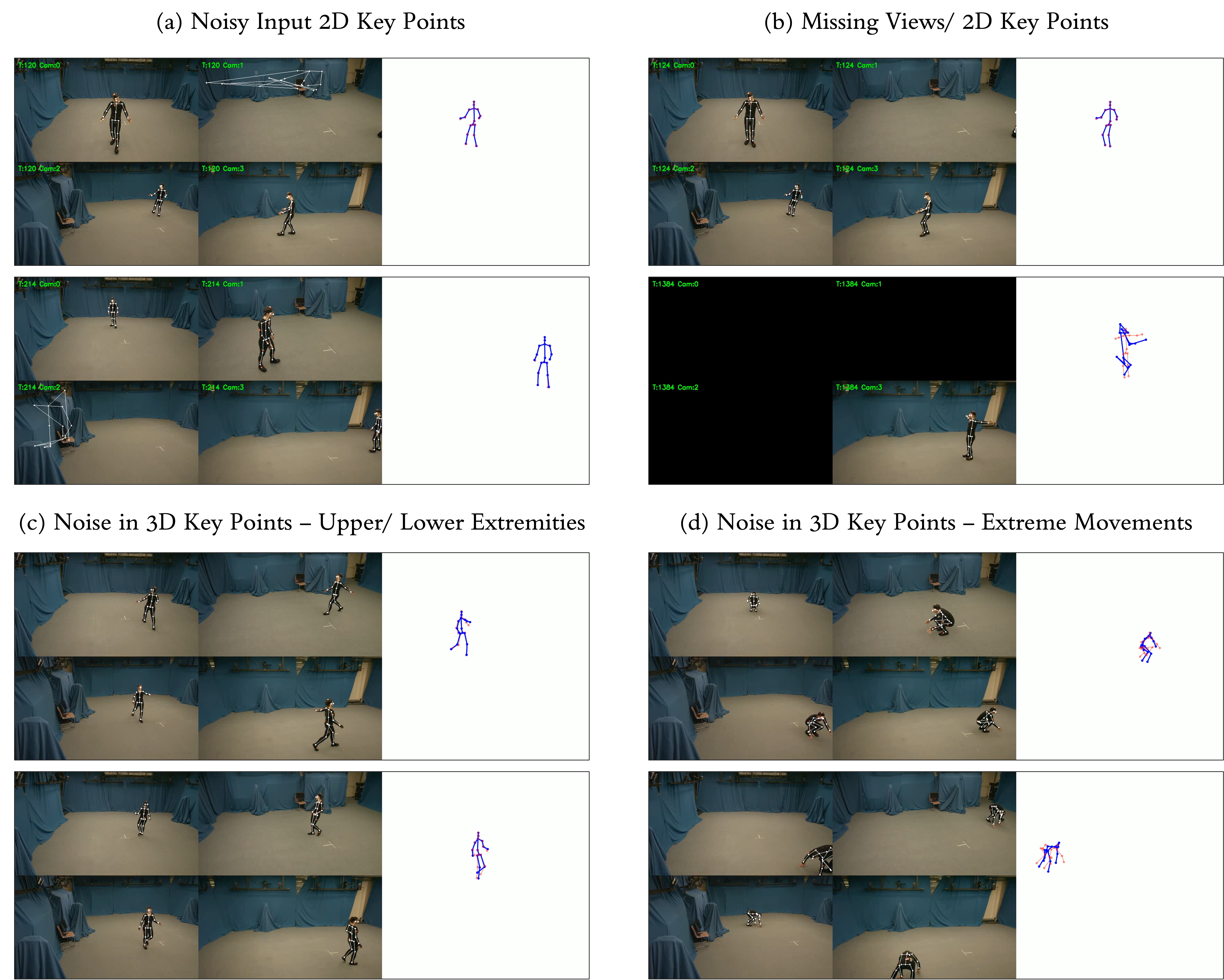}
\caption{Qualitative assessment of STA-TFM under challenging conditions; blue poses represent predictions and red poses represent ground truth. (\textbf{a}) Noisy input 2D keypoints when subjects are partially visible or completely absent from camera views. (\textbf{b}) 3D pose reconstruction from remaining camera views when one or more views are missing or contain incomplete 2D keypoint detections. (\textbf{c}) Noise in predicted 3D keypoints at the extremities, particularly at wrists and arms, compared to stable core body keypoints. (\textbf{d}) Noise in predicted 3D keypoints during extreme movements in freestyle scenarios where rapid motion challenges temporal modeling.\label{fig:qualitative_assessment}}
\end{figure}

Despite these strengths, an increase in noise is observed in the extremities, particularly at the wrists and elbows. As these distal joints typically exhibit higher velocities and larger ranges of motion, they challenge temporal consistency, while core body keypoints remain comparatively stable (Figure~\ref{fig:qualitative_assessment}(c)). Beyond these localized inaccuracies, the model exhibits broader failure cases in extreme scenarios. In the event the majority of camera views lack valid 2D keypoints, the model performance degrades significantly. Complex self-occlusions, particularly in the TotalCapture freestyle scenario, where participants perform rapid atypical movements such as tucking into a ball and jumping, challenge the model's accuracy (Figure~\ref{fig:qualitative_assessment}(d)). These cases highlight the limitations of temporal modeling under highly dynamic motion patterns.

%% file: MDPI_Article_Template/Sections/5_conclusion.tex
\label{sec:conclusion}
This paper presents STA-TFM, a multi-view spatio-temporal transformer for 3D human pose estimation that integrates DSTformer for temporal modeling with a fusion transformer for cross-view aggregation. Unlike traditional approaches that depend on known camera calibration parameters, STA-TFM learns to aggregate multi-view 2D poses to directly predict 3D body poses. This makes STA-TFM more practical for real-world deployments where calibration data may be unavailable or unreliable.

Extensive experiments show that STA-TFM outperforms comparable camera-parameter-free methods. On DHP19, STA-TFM achieves 50.9\% and 49.5\% reductions in P-MPJPE and P-MPJVE compared to MPL, with improvements of 6.7\% and 7.7\% on HAA4D. On TotalCapture, STA-TFM outperforms all compared camera-parameter-free methods, achieving 
15.2\%, 35.7\%, 40.8\%, 50.2\%, and 55.22\% reductions in P-MPJPE compared to MPL \cite{ghasemzadeh2024mpl}, SGraFormer \cite{zhang2024deep}, MTF-Transformer \cite{shuai2022adaptive}, EFMK \cite{zhang2025efmk}, and ESMFormer \cite{zhang2025esmformer}, respectively. The composite loss function $\mathcal{L}_{\text{composite}}$ is shown to enhance temporal consistency, yielding 41.0\% and 2.3\% additional P-MPJVE reductions on DHP19 and HAA4D, respectively, compared to a standard loss function solely based on MPJPE. Qualitative and quantitative results demonstrate tolerance to missing views and noisy keypoints.

To the best of our knowledge, STA-TFM is the first method to leverage frozen pre-trained monocular features (DSTformer) for multi-view 3D HPE.
The limited gains observed from fine-tuning DSTFormer (Section \ref{subsec:ablation_dstformer}) suggest that spatio-temporal representations learned for monocular pose estimation data remain effective when transferred to multi-view 3D human pose estimation.
To address training data scarcity, a data generation pipeline is utilized that simulates $n$ radially distributed cameras with controllable noise parameters, enabling training without expensive multi-camera setups.

 STA-TFM has several limitations. First, the model requires synchronized inputs, which increases computational cost with additional cameras. Second, while STA-TFM operates without explicit geometric priors at inference, the model learns from a specific radial distribution during training; consequently, it operates optimally with similar setups. Furthermore, like other camera-parameter-free methods, STA-TFM supports only single-person inference.  In multi-person scenarios, it requires per-person multi-view keypoints, necessitating keypoint association across views before inference, adding an additional step. Lastly, STA-TFM outputs normalized, root-centered 3D poses, and Procrustes alignment with ground truth  is needed to obtain poses in the world frame of reference. This means that while the predicted poses maintain accurate relative joint positions and can be used to measure body angles and skeletal proportions, they cannot be used to measure real-world distances between joints or track the subject's overall travel distance.

The practical implications of STA-TFM extend across domains where flexible, low-cost motion capture is needed. By only requiring radially distributed cameras with no calibration, STA-TFM enables deployment in environments where traditional motion capture is impractical. In clinical settings, clinicians could capture high-quality 3D pose estimates using low cost smartphones on tripods for quantitative assessment of neuromotor tasks, rehabilitation exercises, and movement disorders. This capability opens the door to generating large, matched datasets of videos and 3D poses for patients performing clinical assessments, data that is currently scarce and expensive to collect. Similar setups could support athletic performance analysis and sports medicine, making 3D pose estimation accessible to practitioners in gyms, on fields, and in rehabilitation facilities.

Although a performance gap remains relative to camera-parameter methods that leverage geometric priors, STA-TFM narrows this gap without requiring camera calibration. Closing the remaining gap would benefit from exploring world-frame pose representations to eliminate the need for Procrustes alignment, extending the fusion mechanism to multi-person scenarios, and optimizing for real-time inference. Establishing standardized benchmarks for camera-parameter-free multi-view HPE would also accelerate progress by enabling fair cross-method comparison across the field.

%% file: MDPI_Article_Template/Sections/appendix.tex
\subsubsection{Performance Against Triangulation Baseline}
\label{app:triangulation}
To assess the impact of using a 2D keypoint detector on the 2D-to-3D lifting process, STA-TFM is evaluated on videos from the BlendMimic3D dataset using 2D keypoints predicted by RTMPose \cite{jiang2023rtmpose}. 
This setup reflects a realistic inference scenario in which the 2D inputs contain inherent detection noise from a 2D HPE model. The resulting 3D poses are compared against linear triangulation \cite{hartley2004multiple}, which estimates 3D keypoints from multi-view 2D observations and corresponding camera parameters. Experiments are conducted across 2-, 3-, and 4-view configurations, as summarized in Table~\ref{tab:2d_pose_effect}.

Triangulation provides a meaningful baseline only when 2D poses are obtained from a keypoint detector rather than ground truth annotations. Under synthetic conditions with perfect camera parameters, triangulation gains an unrealistic advantage because ground truth 2D keypoints together with exact camera parameters allow near deterministic recovery of the original 3D pose when subjects are fully visible and unoccluded across views~\cite{hartley2004multiple}. Although triangulation still benefits from the ideal camera parameters provided in the BlendMimic3D dataset, noise introduced by 2D keypoint detection results in a more realistic and informative performance benchmark.

As shown in Table~\ref{tab:2d_pose_effect}, the P-MPJPE obtained with ground truth 2D poses is 34.2\% lower than that obtained using the RTMPose-predicted 2D keypoints. The increase in P-MPJPE can be attributed to the propagation of errors from the 2D pose detector to the subsequent 3D lifting process. Despite this degradation, STA-TFM consistently outperforms triangulation across all camera configurations in terms of P-PCK and P-MPJVE. Notably, STA-TFM with three cameras achieves lower MPJPE than triangulation with four, reducing setup cost and complexity. STA-TFM also enforces greater temporal coherence and reduces jitter, as demonstrated by lower P-MPJVE values. This highlights STA-TFM’s robustness and generalization when relying solely on noisy 2D detections without requiring camera calibration. 

\begin{table}[H]
%\small % Change table font size
\caption{Performance comparison of STA-TFM and triangulation using both ground truth and RTMPose 2D keypoints on the BlendMimic3D dataset. Lower $\downarrow$ is better; higher $\uparrow$ is better. Best results are bolded.\label{tab:2d_pose_effect}}
%\isPreprints{\centering}{} % Only used for preprints
\begin{tabularx}{\textwidth}{CCCCCC}
\toprule
\textbf{No. Cameras} & \textbf{Model} & \textbf{2D Keypoints Source} & \textbf{P-MPJPE (mm) $\downarrow$} & \textbf{P-PCK (\%) $\uparrow$} & \textbf{P-MPJVE (mm/frame) $\downarrow$}\\
\midrule
\multirow[m]{3}{*}{4} & STA-TFM       & Ground truth              & 61.97           & 85.92           & 2.00\\
\cmidrule(lr){4-6}
                      & STA-TFM       & \multirow[m]{2}{*}{RTMPose} & \textbf{94.17}  & \textbf{70.07}  & \textbf{4.77}\\
                      & Triangulation &                            & 99.33           & 49.10           & 8.62\\
\midrule
\multirow[m]{2}{*}{3} & STA-TFM       & \multirow[m]{2}{*}{RTMPose} & \textbf{98.39}  & \textbf{67.61}  & \textbf{4.86}\\
                      & Triangulation &                            & 99.59           & 48.98           & 9.46\\
\midrule
\multirow[m]{2}{*}{2} & STA-TFM       & \multirow[m]{2}{*}{RTMPose} & 108.02          & \textbf{60.39}  & \textbf{4.45}\\
                      & Triangulation &                            & \textbf{101.09} & 47.84           & 10.78\\
\bottomrule
\end{tabularx}
\end{table}